\documentclass[twoside,english,3p]{elsarticle}
\usepackage[T1]{fontenc}
\usepackage{geometry}
\usepackage{amssymb}
\usepackage{amsmath}

\usepackage{amsthm}
\usepackage{stmaryrd}
\usepackage{graphicx}
\usepackage{booktabs}
\usepackage{multirow}
\usepackage{makecell}
\usepackage{xurl}
\usepackage{esint}
\usepackage{algorithm}
\usepackage{algpseudocode}
\usepackage{rotating}
\usepackage{adjustbox}
\usepackage{nomencl}
\usepackage{multicol}
\makenomenclature
\makeatletter
\theoremstyle{plain}

\theoremstyle{boldremark} 

\ifx\proof\undefined

\providecommand{\proofname}{Proof}
\fi

\renewenvironment{proof}[1][\proofname]{%
\par\pushQED{\qed}\normalfont%
\topsep6\p@\@plus6\p@\relax
\trivlist\item[\hskip\labelsep\bfseries#1\@addpunct{.}]%
\ignorespaces
}{%
\popQED\endtrivlist\@endpefalse
}

\journal{Elsevier}

\usepackage{hyperref}
\hypersetup{colorlinks = true, allcolors = blue}

\usepackage[nameinlink]{cleveref}

\crefname{figure}{Fig.}{Figs.}
\crefformat{equation}{Eq.~#2(#1)#3}
\crefformat{section}{Section~#2#1#3}
\AtBeginDocument{%
\let\citet\cite
}

\usepackage[labelfont=bf]{caption}
\@ifundefined{showcaptionsetup}{}{%
\PassOptionsToPackage{caption=false}{subfig}}
\usepackage{subfig}
\makeatother

\usepackage{babel}
\providecommand{\remarkname}{Remark}
\providecommand{\theoremname}{Theorem}

\begin{document}

\begin{frontmatter}{}
	
	\title{Transfer Learning in Physics-Informed Neural Networks: Full Fine-Tuning, Lightweight Fine-Tuning, and Low-Rank Adaptation}

	\author[rvt,rvt3]{Yizheng Wang}

\ead{wang-yz19@tsinghua.org.cn}

\author[rvt,rvt4,rvt5]{Jinshuai Bai}

\author[rvt6]{Mohammad Sadegh Eshaghi}

\author[rvt3]{Cosmin Anitescu}

\author[rvt6]{Xiaoying Zhuang}

\author[rvt3]{Timon Rabczuk}

\author[rvt]{Yinghua Liu\corref{cor1}}

\ead{yhliu@mail.tsinghua.edu.cn}
\cortext[cor1]{Corresponding author}
\address[rvt]{Department of Engineering Mechanics, Tsinghua University, Beijing 100084, China}

\address[rvt3]{Institute of Structural Mechanics, Bauhaus-Universit\"{a}t Weimar, Marienstr. 15, D-99423 Weimar, Germany}	

\address[rvt6]{ Institute of Photonics, Department of Mathematics and Physics, Leibniz University Hannover, Germany}

\address[rvt2]{Drilling Mechanical Department, CNPC Engineering Technology RD Company Limited, Beijing 102206, China}

\address[rvt4]{School of Mechanical, Medical and Process Engineering, Queensland University of Technology, Brisbane, QLD 4000, Australia}

\address[rvt5]{ARC Industrial Transformation Training Centre—Joint Biomechanics, Queensland University of Technology, Brisbane, QLD 4000, Australia}

\begin{abstract}

AI for PDEs has garnered significant attention, particularly Physics-Informed Neural Networks (PINNs). However, PINNs are typically limited to solving specific problems, and any changes in problem conditions necessitate retraining. Therefore, we explore the generalization capability of transfer learning in the strong and energy form of PINNs across different boundary conditions, materials, and geometries. The transfer learning methods we employ include full finetuning, lightweight finetuning, and Low-Rank Adaptation (LoRA). 
The results demonstrate that full finetuning and LoRA can significantly improve convergence speed while providing a slight enhancement in accuracy.

\end{abstract}

\printnomenclature

\begin{graphicalabstract}
	\includegraphics[scale=0.75]{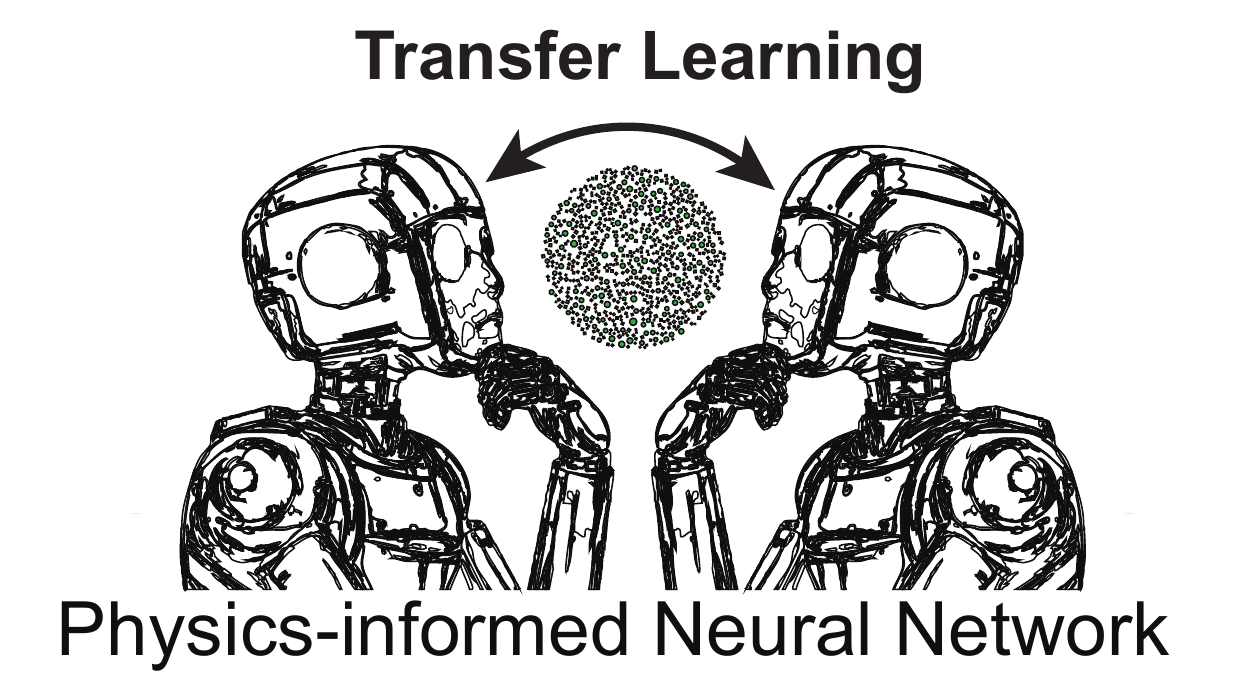}
\end{graphicalabstract}

\begin{keyword}
	PINNs \sep  Computational mechanics \sep Transfer learning  \sep
	AI for PDEs \sep AI for science 
\end{keyword}

\end{frontmatter}{}

\section{Introduction}
Numerous physical phenomena are modeled using PDEs \citet{loss_is_minimum_potential_energy}. Solving PDEs is key to understanding these phenomena, making it one of the most critical tasks in computational mathematics, physics, and mechanics \citet{PINN_review}. However, traditional PDEs solvers require re-solving the problem whenever boundary conditions, material distributions, or geometries change \citet{wang2021learning}.

Recently, AI for PDEs \citet{wang2024artificial,yizheng2024ai}, one of the important directions of AI for science, refers to a class of algorithms that use deep learning to solve PDEs. There are three main approaches in AI for PDEs. The first is Physics-Informed Neural Networks (PINNs) \citet{PINN_original_paper}, which are based on physical equations. The second is data-driven operator learning \citet{DeepOnet,li2020fourier}. The third is hybrid methods that combine data and physics, such as Physics-informed Neural Operator (PINO) \citet{li2024physics} and Variational Physics-informed Neural Operator (VINO) \citet{eshaghi2024variational}. PINNs are a core component of AI for PDEs, but they are limited to solving specific problems. Any changes in boundary conditions, material distributions, or geometries necessitate retraining \citet{yang2023context,desai2021one,gao2022svd}.

Transfer learning refers to fine-tuning a pre-trained model for a new related task, often requiring fewer iterations and less data for the new task \citet{zhuang2020comprehensive}. Therefore, exploring transfer learning in PINNs for different boundary conditions \citet{xu2023transfer}, material distributions \citet{guo2022analysis}, and geometries \citet{chakraborty2022domain} is of great importance. The advantage of transfer learning in PINNs lies in iterative algorithms, as they can inherit parameters from previous training, thereby reducing the number of iterations required for new but similar tasks \citet{wang2024artificial}. Currently, there are two main methods for applying transfer learning in PINNs: The first is full finetuning, where the model parameters are fully inherited from the similiar task and then the model is re-trained \citet{chen2021transfer,PINN_solid_mechanics}. The second is lightweight finetuning, where the initial layers of the neural network are frozen, and only the latter layers are trained to achieve less literations to converge \citet{goswami2020transfer,chakraborty2021transfer}. Recently, LoRA (Low-Rank Adaptation of Large Language Models) \citet{hu2021lora} has been introduced, which uses low-rank approximation of trainable parameters to enable rapid fine-tuning, significantly reducing computational costs. LoRA offers a more generalized and flexible approach compared to full and lightweight finetuning. \citet{majumdar2023hyperlora,cho2023hypernetwork} have applied LoRA to PINNs. However, there has been no systematic study on the performance of different transfer learning methods in both the strong and energy forms of PINNs, particularly for generalization across boundary conditions, geometries, and material distributions.
Given that transfer learning is a pivotal technique for enhancing the adaptability of PINNs across varying scenarios, a comprehensive analysis is essential to understand performance of transfer learning in PINNs.

As a result, we systematically tested different transfer learning methods in PINNs for varying boundary conditions, geometries, and material distributions. The transfer learning methods include full finetuning, lightweight finetuning, and LoRA. Our results show that full finetuning and LoRA can significantly improve convergence speed while providing a slight enhancement in accuracy across most scenarios.
These findings underscore the potential of transfer learning techniques to optimize the training efficiency and performance of PINNs, particularly in complex physical systems with varying boundary conditions, geometries, and material distributions.

The outline of the paper is as follows. \Cref{sec:Preparatory-knowledge} introduces PINNs, divided into strong and energy form. \Cref{sec:Method:-Transfer-learning} presents different transfer learning methods. \Cref{sec:Results} details numerical experiments, divided into three parts:
\begin{enumerate}
	\item Transfer learning for different boundary conditions: Solving incompressible fluid dynamics Navier-Stokes (NS) equations with varying boundary conditions using the strong form of PINNs.
	\item Transfer learning for different materials: Solving heterogeneous problems with functionally graded materials of varying elastic moduli using the energy form of PINNs.
	\item Transfer learning for different geometries: Solving the square plate with a hole problem with varying geometries using the energy form of PINNs.
\end{enumerate}
Finally, \Cref{sec:Dicussion} and \Cref{sec:Conclusion} summarize the characteristics of different transfer learning methods in PINNs and suggest future research directions for transfer learning in PINNs.

\section{Preparatory knowledge\label{sec:Preparatory-knowledge}}

Before introducing transfer learning in PINNs, we need to understand two important forms of PINNs: the strong form of PINNs \citet{PINN_original_paper} and the energy form \citet{loss_is_minimum_potential_energy}. The essential difference between these two forms lies in the mathematical representation of the PDEs. The strong form describes the equations from a differential element perspective, while the energy form uses variational principles to describe the physical system from the overall energy perspective \citet{the_foundation_of_solid_mechanics_feng}. It is important to note that while the PDEs may have different forms, they solve the same physical problem. The reason for studying different forms of PDEs lies in the fact that, although they represent the same physical problem, their algorithmic accuracy and efficiency differ \citet{wang2025physics}. Therefore, studying different forms of PINNs is crucial for AI for PDEs \citet{yizheng2024ai,wang2024artificial}, which is one of the important aspect in AI for Science \citet{zhang2023artificial}. Below, we introduce both the strong form and the energy form.

\subsection{PINNs: Strong Form}

\begin{figure}
	\begin{centering}
		\includegraphics{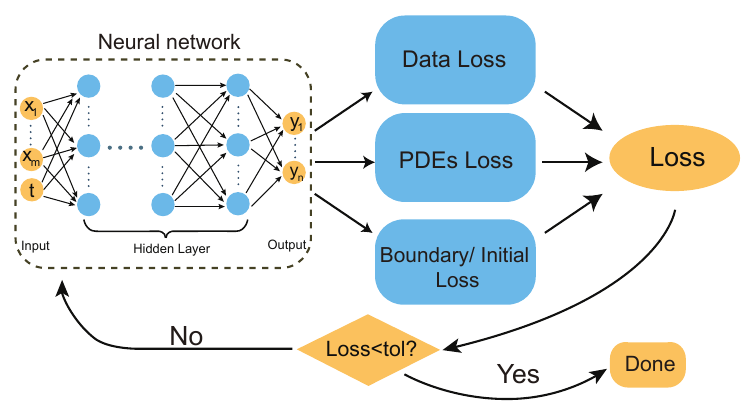}
		\par\end{centering}
	\caption{The strong form of Physics-informed neural networks (PINNs). The inputs $x_{1}$, $x_{2}$, $\cdots$, $t$ are typically spatial coordinates, while the outputs $y_{1}$, $\cdots$, $y_{n}$ are the network's outputs, usually representing the variables of interest.\label{fig:Physics-informed-neural-networks}}
\end{figure}

We begin by discussing the PDEs for boundary value problems, considering the following equations:
\begin{equation}
	\begin{cases}
		\boldsymbol{P}(\boldsymbol{u}(\boldsymbol{x}))=\boldsymbol{f}(\boldsymbol{x}) & \boldsymbol{x}\in\Omega\\
		\boldsymbol{B}(\boldsymbol{u}(\boldsymbol{x}))=\boldsymbol{g}(\boldsymbol{x}) & \boldsymbol{x}\in\Gamma
	\end{cases}\label{eq:original_form},
\end{equation}
where $\boldsymbol{P}$ and $\boldsymbol{B}$ are the domain and boundary operators, respectively, and $\Omega$ and $\Gamma$ represent the domain and boundary.
\(\boldsymbol{u}(\boldsymbol{x})\) denote the field of interest that needs to be solved, where \(\boldsymbol{x}\) represents the spatial or temporal coordinates.

We use the weighted residual method to transform these equations into their weighted residual form:
\begin{equation}
	\begin{cases}
		\int_{\Omega}[\boldsymbol{P}(\boldsymbol{u}(\boldsymbol{x}))-\boldsymbol{f}(\boldsymbol{x})]\cdot\boldsymbol{w}(\boldsymbol{x})d\Omega=0 & \boldsymbol{x}\in\Omega\\
		\int_{\Gamma}[\boldsymbol{B}(\boldsymbol{u}(\boldsymbol{x}))-\boldsymbol{g}(\boldsymbol{x})]\cdot\boldsymbol{w}(\boldsymbol{x})d\Gamma=0 & \boldsymbol{x}\in\Gamma
	\end{cases}\label{eq:original_form_weighted},
\end{equation}
where $\boldsymbol{w}(\boldsymbol{x})$ is the weight function.  \Cref{eq:original_form} and \Cref{eq:original_form_weighted} are equivalent if $\boldsymbol{w}(\boldsymbol{x})$ is arbitrary. For numerical convenience, we often predefine the form of $\boldsymbol{w}(\boldsymbol{x})$ and obtain the residual form of the PDEs:
\begin{equation}
	\boldsymbol{w}(\boldsymbol{x})=\begin{cases}
		\boldsymbol{P}(\boldsymbol{u}(\boldsymbol{x}))-\boldsymbol{f}(\boldsymbol{x}) & \boldsymbol{x}\in\Omega\\
		\boldsymbol{B}(\boldsymbol{u}(\boldsymbol{x}))-\boldsymbol{g}(\boldsymbol{x}) & \boldsymbol{x}\in\Gamma
	\end{cases}.\label{eq:strong_form_weight}
\end{equation}
As a result, \Cref{eq:original_form_weighted} is transformed into:
\begin{equation}
	\begin{cases}
		\int_{\Omega}[\boldsymbol{P}(\boldsymbol{u}(\boldsymbol{x}))-\boldsymbol{f}(\boldsymbol{x})]\cdot[\boldsymbol{P}(\boldsymbol{u}(\boldsymbol{x}))-\boldsymbol{f}(\boldsymbol{x})]d\Omega=0 & \boldsymbol{x}\in\Omega\\
		\int_{\Gamma}[\boldsymbol{B}(\boldsymbol{u}(\boldsymbol{x}))-\boldsymbol{g}(\boldsymbol{x})]\cdot[\boldsymbol{B}(\boldsymbol{u}(\boldsymbol{x}))-\boldsymbol{g}(\boldsymbol{x})]d\Gamma=0 & \boldsymbol{x}\in\Gamma
	\end{cases}\label{eq:original_form_delta}.
\end{equation}

Next, we approximate these integrals shown in \Cref{eq:original_form_delta} numerically, leading to the strong form of PINNs:
\begin{equation}
\mathcal{L}_{PINNs}=\frac{\lambda_{r}}{N_{r}}\sum_{i=1}^{N_{r}}||\boldsymbol{P}(\boldsymbol{u}(\boldsymbol{x}_{i};\boldsymbol{\theta}))-\boldsymbol{f}(\boldsymbol{x}_{i})||_{2}^{2}+\frac{\lambda_{b}}{N_{b}}\sum_{i=1}^{N_{b}}||\boldsymbol{B}(\boldsymbol{u}(\boldsymbol{x}_{i};\boldsymbol{\theta}))-\boldsymbol{g}(\boldsymbol{x}_{i})||_{2}^{2}.
\end{equation}
,where $||\cdot||_{2}$ is the $L_2$ norm.

We optimize the above loss function to obtain the neural network approximation of the field variable $\boldsymbol{u}(\boldsymbol{x};\boldsymbol{\theta})$, where \(\boldsymbol{\theta}\) is the parameters of the neural network:
\begin{equation}
\boldsymbol{u}(\boldsymbol{x};\boldsymbol{\theta})=\underset{\boldsymbol{\theta}}{\arg\min}\mathcal{L}_{PINNs}
\end{equation}

Thus, mathematically, the strong form of PINNs \citep{PINN_original_paper} is essentially a weighted residual form with the approximation of the function $\boldsymbol{u}(\boldsymbol{x})$ using a neural network, and the weight function given by \Cref{eq:strong_form_weight}.

\subsection{PINNs: Energy Form}

\begin{figure}
	\begin{centering}
		\includegraphics[scale = 0.6]{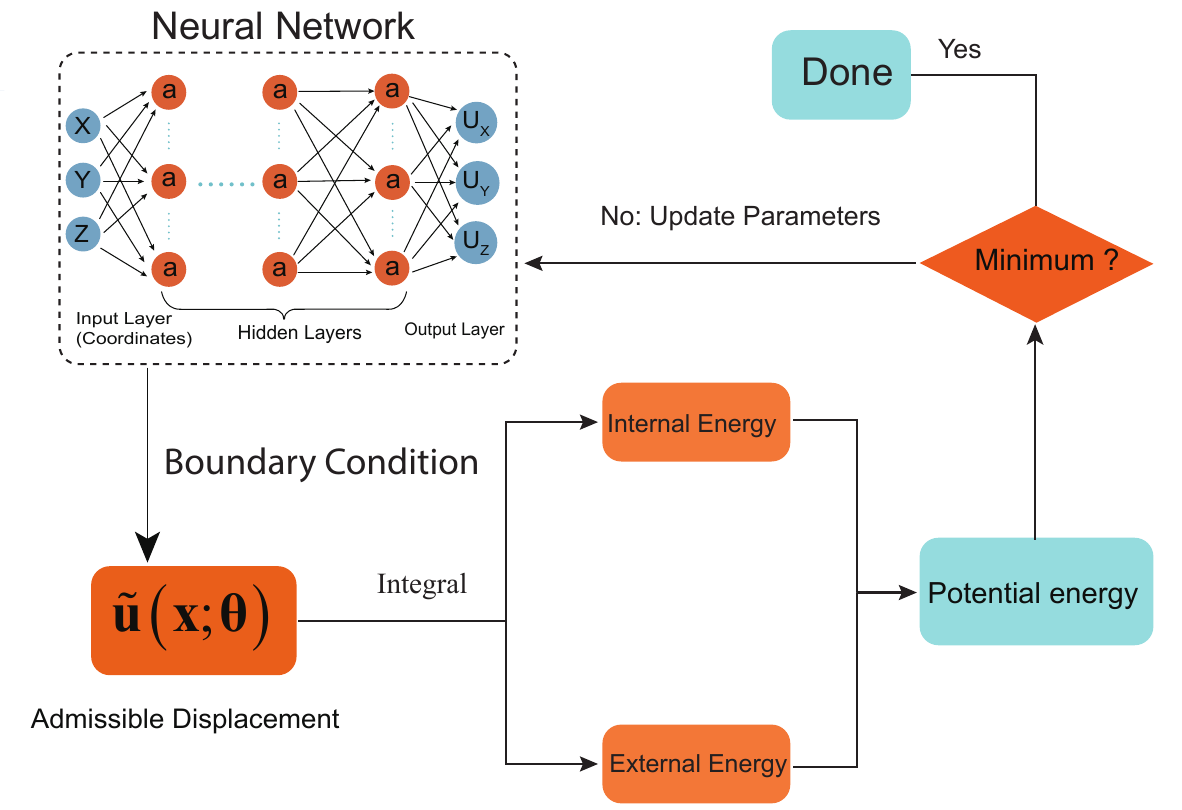}
		\par\end{centering}
	\caption{The energy form of Physics-informed neural networks (PINNs). The inputs $x$, $y$, $z$ are typically spatial coordinates, while the outputs $U_{x}$, $U_{y}$, and $U_{z}$ represent the displacement field $\boldsymbol{u}$. The admissible displacement $\tilde{\boldsymbol{u}}(\boldsymbol{x};\boldsymbol{\theta})$ is the displacement field $\boldsymbol{u}$ that satisfies the essential boundary conditions in advance.\label{fig:DEM}}
\end{figure}

We introduce the energy form of the Physics-Informed Neural Network (DEM: Deep Energy Method) \citep{loss_is_minimum_potential_energy}. 
We consider $\delta\boldsymbol{u}$ as the weight function $\boldsymbol{w}(\boldsymbol{x})$ in \Cref{eq:original_form_weighted}, 
which leads to the Galerkin form. \Cref{eq:original_form_weighted} can be written as:
\begin{align}
	\int_{\Omega}[\boldsymbol{P}(\boldsymbol{u}(\boldsymbol{x}))-\boldsymbol{f}(\boldsymbol{x})]\cdot\delta\boldsymbol{u}\,d\Omega=0 & ,\boldsymbol{x}\in\Omega.\label{eq:galerkin_form}
\end{align}

For simplicity, we consider a specific Poisson equation to illustrate this:
\begin{equation}
	\begin{cases}
		-\triangle(u(\boldsymbol{x}))=f(\boldsymbol{x}) & \boldsymbol{x}\in\Omega\\
		u(\boldsymbol{x})=\bar{u}(\boldsymbol{x}) & \boldsymbol{x}\in\Gamma^{u}\\
		\frac{\partial u(\boldsymbol{x})}{\partial\boldsymbol{n}}=\bar{t}(\boldsymbol{x}) & \boldsymbol{x}\in\Gamma^{t}
	\end{cases}.\label{eq:poisson_equation}
\end{equation}
where $\Gamma^{u}$ and $\Gamma^{t}$ are the Dirichlet and Neumann boundary conditions, respectively. 
For the Poisson equation, the Galerkin form of \Cref{eq:galerkin_form} can be expressed as:
\begin{align}
	\int_{\Omega}[-\triangle(u(\boldsymbol{x}))-f(\boldsymbol{x})]\cdot\delta u\,d\Omega=0 & ,\boldsymbol{x}\in\Omega.\label{eq:galerkin_form_poisson}
\end{align}

Using the Gaussian integration formula, we can transform the above equation to:
\begin{equation}
	\int_{\Omega}(-u_{,ii}-f)\delta u\,d\Omega=\int_{\Omega}u_{,i}\delta u_{,i}\,d\Omega-\int_{\Gamma}u_{,i}n_{i}\delta u\,d\Gamma-\int_{\Omega}f\delta u\,d\Omega=0.\label{eq:gaussian_poisson}
\end{equation}
By incorporating the boundary conditions from \Cref{eq:poisson_equation} into \Cref{eq:gaussian_poisson}, 
we obtain the Galerkin weak form:
\begin{equation}
	\int_{\Omega}(-u_{,ii}-f)\delta u\,d\Omega=\int_{\Omega}u_{,i}\delta u_{,i}\,d\Omega-\int_{\Gamma^{t}}\bar{t}\delta u\,d\Gamma-\int_{\Omega}f\delta u\,d\Omega=0.\label{eq:weak_form}
\end{equation}
Since $u(\boldsymbol{x})$ is given on $\Gamma^{u}$, the corresponding variation $\delta\boldsymbol{u}=0$ 
on $\Gamma^{u}$. Here, we observe an interesting phenomenon: we must satisfy $u(\boldsymbol{x})=\bar{u}(\boldsymbol{x})$ 
on $\Gamma^{u}$ in advance, which involves constructing an admissible function. This is crucial for 
DEM. Additionally, \Cref{eq:weak_form} includes the domain PDEs and the boundary conditions on $\Gamma^{t}$. 
Therefore, solving \Cref{eq:poisson_equation} is equivalent to solving \Cref{eq:weak_form}.

We can further use the variational principle to write \Cref{eq:weak_form} as:
\begin{equation}
	\delta\mathcal{L}=\int_{\Omega}u_{,i}\delta u_{,i}\,d\Omega-\int_{\Gamma^{t}}\bar{t}\delta u\,d\Gamma-\int_{\Omega}f\delta u\,d\Omega.\label{eq:first_vari}
\end{equation}
\begin{equation}
	\mathcal{L}=\frac{1}{2}\int_{\Omega}u_{,i}u_{,i}\,d\Omega-\int_{\Gamma^{t}}\bar{t}u\,d\Gamma-\int_{\Omega}fu\,d\Omega.\label{eq:energy}
\end{equation}

$\mathcal{L}$ represents the potential energy, and we can observe that $\delta^{2}\mathcal{L}>0$ (excluding 
zero solutions), indicating that we solve for $u(\boldsymbol{x})$ by minimizing the energy:
\begin{equation}
	u(\boldsymbol{x})=\underset{u}{\arg\min}\,\mathcal{L}.\label{eq:minimum_energy}
\end{equation}
The essence of DEM is to approximate $u(\boldsymbol{x})$ using a neural network $u(\boldsymbol{x};\boldsymbol{\theta})$, 
and then optimize \Cref{eq:minimum_energy}:
\begin{equation}
	u(\boldsymbol{x};\boldsymbol{\theta})=\underset{\boldsymbol{\theta}}{\arg\min}\,\mathcal{L}_{DEM}=\underset{u}{\arg\min}\left\{\frac{1}{2}\int_{\Omega}u(\boldsymbol{x};\boldsymbol{\theta})_{,i}u(\boldsymbol{x};\boldsymbol{\theta})_{,i}\,d\Omega-\int_{\Gamma^{t}}\bar{t}u(\boldsymbol{x};\boldsymbol{\theta})\,d\Gamma-\int_{\Omega}fu(\boldsymbol{x};\boldsymbol{\theta})\,d\Omega\right\}.\label{eq:DEM}
\end{equation}
Therefore, the core of DEM lies in the integration of the domain energy and boundary energy, as well 
as the construction of the admissible function. Integration strategies can use numerical analysis methods, 
such as simple Monte Carlo integration or more accurate methods like Gaussian integration or Simpson's 
Rule.

Here, we emphasize the construction of the admissible function. We use the concept of a distance network 
for this purpose:
\begin{equation}
	u(\boldsymbol{x})=u_{p}(\boldsymbol{x};\boldsymbol{\theta}_{p})+D(\boldsymbol{x})\cdot u_{g}(\boldsymbol{x};\boldsymbol{\theta}_{g}),\label{eq:admissible}
\end{equation}
where $u_{p}(\boldsymbol{x};\boldsymbol{\theta}_{p})$ is the particular solution network that fits the 
Dirichlet boundary condition, such that it outputs $\bar{u}(\boldsymbol{x})$ when the input points are 
on $\Gamma^{u}$, and outputs any value elsewhere. The parameters $\boldsymbol{\theta}_{p}$ are optimized 
by:
\begin{equation}
\boldsymbol{\theta}_{p}=\underset{\boldsymbol{\theta}_{p}}{\arg\min}\,\text{MSE}(u_{p}(\boldsymbol{x};\boldsymbol{\theta}_{p}),\bar{u}(\boldsymbol{x})),\quad\boldsymbol{x}\in\Gamma^{u}.
\end{equation}
$D(\boldsymbol{x})$ is the distance network, which we approximate using radial basis functions \citep{wang2022cenn}. 
Other fitting functions can also be used. The effect is to output the minimum distance to the Dirichlet 
boundary:
\begin{equation}
	D(\boldsymbol{x})=\min_{\boldsymbol{y}\in\Gamma^{u}}\sqrt{(\boldsymbol{x}-\boldsymbol{y})\cdot(\boldsymbol{x}-\boldsymbol{y})}.
\end{equation}
$u_{g}(\boldsymbol{x};\boldsymbol{\theta}_{g})$ is a standard neural network. When using the minimum 
potential energy principle, we only optimize $\boldsymbol{\theta}_{g}$:
\begin{equation}
	u(\boldsymbol{x};\boldsymbol{\theta}_{p},\boldsymbol{\theta}_{g})=\underset{\boldsymbol{\theta}_{g}}{\arg\min}\,\mathcal{L}_{DEM}.
\end{equation}

\section{Method: Transfer learning in PINNs\label{sec:Method:-Transfer-learning}}

Transfer learning has been a hot topic in deep learning research. In this manuscript, we conduct a systematic evaluation of transfer learning in the strong and energy forms of PINNs.

\begin{figure}
	\begin{centering}
		\includegraphics[scale = 0.6]{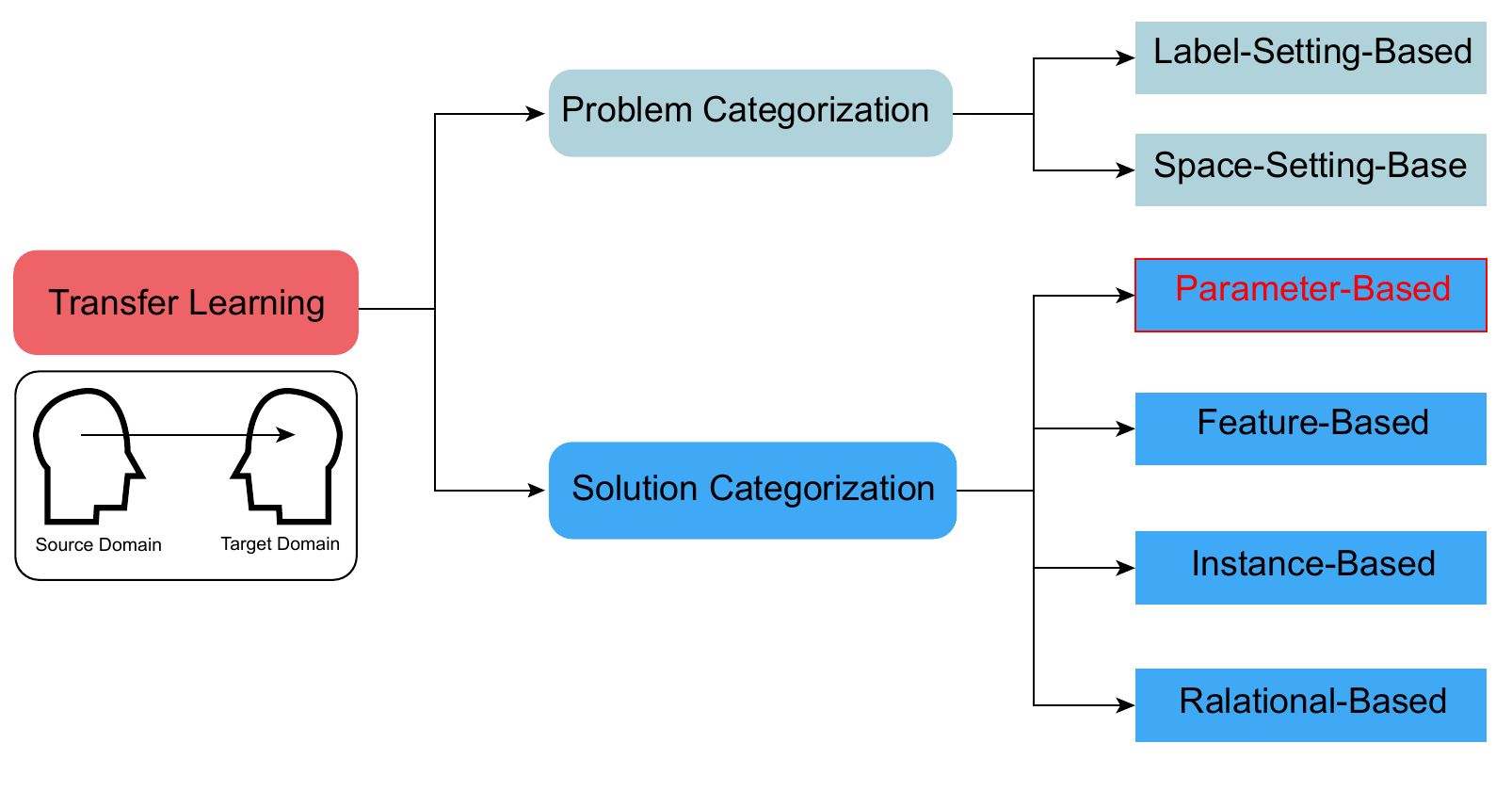}
		\par\end{centering}
	\caption{Categories of Transfer Learning \citet{zhuang2020comprehensive}\label{fig:transfer_learning_categorization}}
\end{figure}

Given that there are various transfer learning methods, as shown in \Cref{fig:transfer_learning_categorization}, we focus on the classical Parameter-based transfer learning and evaluate its performance in PINNs. The source domain refers to the pre-trained source problem, while the target domain refers to the target problem. The core of transfer learning lies in leveraging the parameters pre-trained on the source domain to achieve faster convergence on the target domain.
Below, we introduce three common approaches for Parameter-based transfer learning.

\subsection{Full Fine-Tuning}

The most common method for adapting pretrained models to downstream tasks is full fine-tuning, where all model parameters are fine-tuned \citet{he2021towards}. However, this results in a full copy of fine-tuned model parameters for each task, which becomes prohibitively expensive when the models have a large number of trainable parameters \citet{radford2019language}. Full fine-tuning is easy to understand, as shown in \Cref{fig:Parameter-based-transfer-learn}a. The core of full fine-tuning is to initialize the neural network parameters with those from the old task. Then, on the new task, the parameters from the old task are fine-tuned:
\begin{equation}
\boldsymbol{\theta}^{new} = \arg\min_{\boldsymbol{\theta}} \mathcal{L}(D^{new}; \boldsymbol{\theta}^{old}),
\end{equation}
where $\boldsymbol{\theta}^{old}$ is the parameter set after the old task is completed, and $\boldsymbol{\theta}^{new}$ is the result of optimizing the loss function $\mathcal{L}$ on the new dataset $D^{new}$.

\subsection{Lightweight Fine-Tuning}

Lightweight fine-tuning freezes most of the pretrained parameters and only tune a smaller subset of parameters \citet{he2021towards}, as shown in \Cref{fig:Parameter-based-transfer-learn}b. The key question is how to decide which subset of pretrained parameters to tune.

In the context of AI for PDEs, a common method for lightweight fine-tuning is to freeze the early layers of the neural network and only train the later layers to adapt to the new task \citet{goswami2020transfer, chakraborty2021transfer, xu2023transfer}. The motivation for this approach is that shallow layers typically extract basic features, while retraining deeper layers enables the transfer of abstract knowledge. Due to the significant reduction in trainable parameters, the computational costs are substantially reduced.

\subsection{LoRA: Low-Rank Adaptation}

\begin{figure}
	\begin{centering}
		\includegraphics[scale = 0.6]{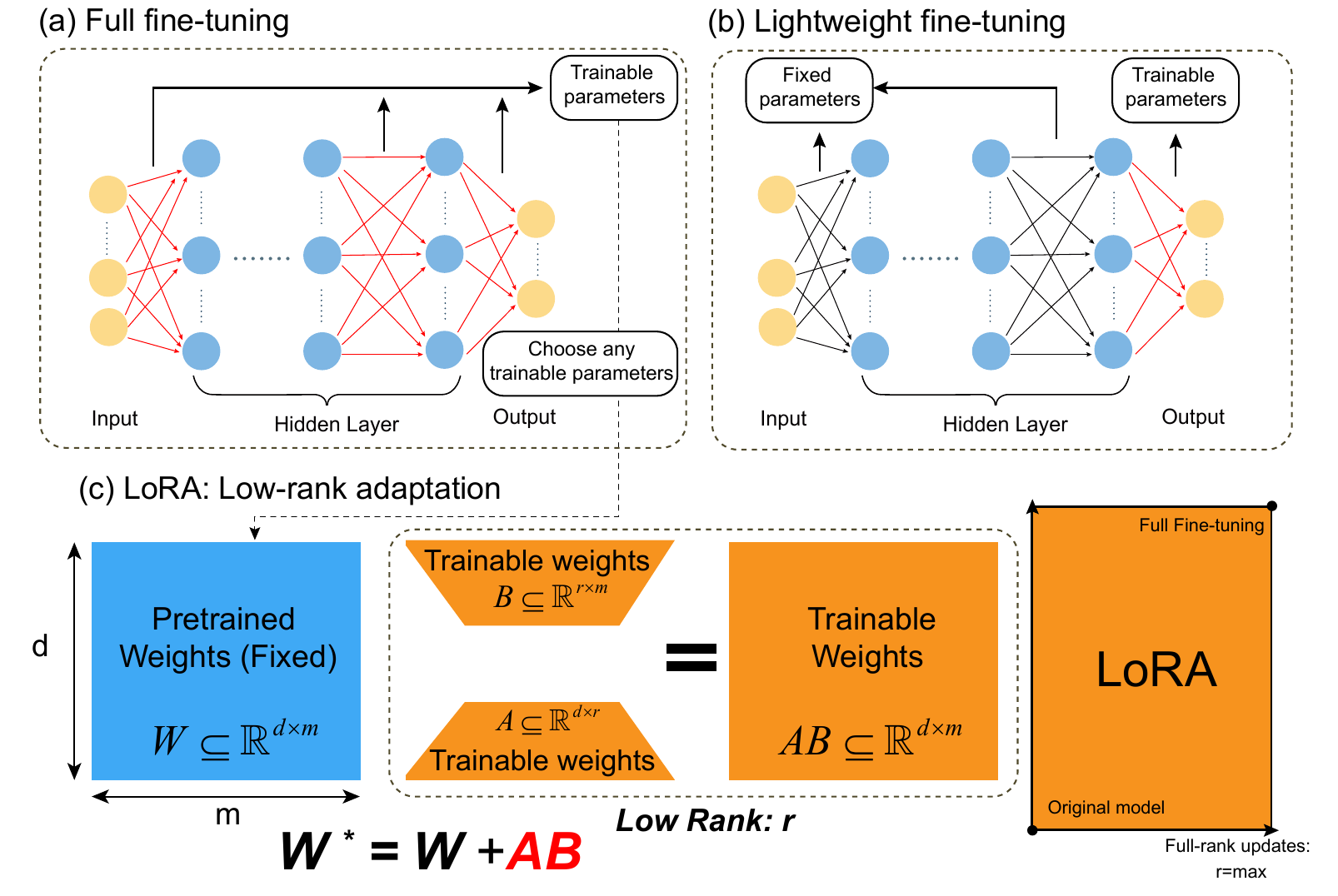}
		\par\end{centering}
	\caption{Three common methods for Parameter-based transfer learning: (a) Full fine-tuning: fine-tune all parameters of the model, with the red arrows indicating the parameters that need fine-tuning. (b) Lightweight fine-tuning: fine-tune a subset of the model’s parameters, with the red arrows indicating the parameters that need fine-tuning. (c) LoRA: the blue matrix $\boldsymbol{W}$ represents fixed parameters (pretrained parameters from the previous task). The yellow matrices are the trainable parameters $\boldsymbol{A}$ and $\boldsymbol{B}$, where $\boldsymbol{AB}$ is a low-rank matrix with rank $r$. During training on the new dataset, only $\boldsymbol{A}$ and $\boldsymbol{B}$ are adjusted. $\boldsymbol{W}^{*} = \boldsymbol{W} + \boldsymbol{A} \boldsymbol{B}$ represents the neural network weights during testing.\label{fig:Parameter-based-transfer-learn}}
\end{figure}

LoRA (Low-Rank Adaptation) \citet{hu2021lora} proposes multiplying low-rank matrices $\boldsymbol{AB}$ (where $\boldsymbol{A} \subseteq \mathbb{R}^{d \times r}$, $\boldsymbol{B} \subseteq \mathbb{R}^{r \times m}$), and then adding them to the pretrained parameters $\boldsymbol{W} \subseteq \mathbb{R}^{d \times m}$, as shown in \Cref{fig:Parameter-based-transfer-learn}c. The parameter computation for LoRA is as follows:
\begin{equation}
\boldsymbol{W}^{*} = \boldsymbol{W} + \alpha \boldsymbol{A} \boldsymbol{B}\label{eq:lora_computation},
\end{equation}
where $r$ is the rank and hyperparameter in LoRA. In general, $r \ll \min(d, m)$, and we discussed the method for determining $r$ in \Cref{subsec:The-rank-in-LoRA}.

It is important to note that $\boldsymbol{W}$ represents the fixed pretrained parameters from the old task (source domain). Only the product $\boldsymbol{A} \boldsymbol{B}$ is trained in the target domain, with the number of trainable parameters being $r \times (d + m)$. The number of trainable parameters in $\boldsymbol{A} \boldsymbol{B}$ is much smaller than the total number of parameters in $\boldsymbol{W}$, which is $d \times m$. The scaling factor $\alpha$ is used to control the weight between the pretrained $\boldsymbol{W}$ and the LoRA-trained parameters $\boldsymbol{A} \boldsymbol{B}$, with a default value of 1.
$\boldsymbol{A}$ and $\boldsymbol{B}$ are initialized using a Gaussian distribution. Other initialization methods for $\boldsymbol{A}$ and $\boldsymbol{B}$ are also possible. In this work, we employ a Gaussian distribution with a mean of $0$ and a standard deviation of $0.02$.

In essence, LoRA can be seen as a more flexible version of lightweight fine-tuning. Additionally, full fine-tuning can be considered a limiting case of LoRA, when $r = \min(d, m)$. The core idea of LoRA is to decompose part of the weight matrices in the model into the product of two low-rank matrices, thereby reducing the number of parameters that need to be trained.

\section{Results\label{sec:Results}}
In this section, we evaluate the performance of the three transfer learning schemes introduced in \Cref{sec:Method:-Transfer-learning} within the context of PINNs. We conduct systematic and in-depth tests to examine the transfer effects across different boundary conditions, material distributions, and geometric configurations.
\subsection{Transfer learning for different boundary conditions\label{subsec:Transfer_boundary}}

Since PINNs can only solve for a specific case, this means that once the boundary or initial conditions change, PINNs must be retrained. In this section, we validate the performance of transfer learning in the strong form of PINNs when the boundary and initial conditions change. 
We perform the validation by the strong form of PINNs on the incompressible fluid dynamics Navier-Stokes (NS) equations, which are a 2D unsteady and nonlinear PDEs system:
\begin{equation}
	\begin{alignedat}{1}
		u_{i,i} & =0 \\
		\frac{\partial u_{i}}{\partial t} + u_{j} u_{i,j} & = -p_{,i} + \frac{1}{Re} u_{i,jj}
	\end{alignedat}
	\label{eq:NS_original},
\end{equation}
where the pressure \( p \) and velocity field \( \boldsymbol{u} \) (where \( u \) represents the velocity field in the \( x \)-direction, and \( v \) represents the velocity field in the \( y \)-direction) are strongly coupled. 
Noting that \Cref{eq:NS_original} cannot be solved by the energy form of PINNs (for reasons detailed in \citet{wang2025physics}), so we focus here on validating the strong form of PINNs.

Although solving \Cref{eq:NS_original} with the SIMPLE algorithm \citep{patankar1983calculation}, a widely recognized and effective method for solving Navier-Stokes equations, is successful, the absence of the SIMPLE method typically results in a large, irregular, and stiff sparse linear system after discretizing \Cref{eq:NS_original}. Solving the poorly-conditioned linear system of \Cref{eq:NS_original} is computationally expensive and difficult to converge. Therefore, fully coupled solution methods are rarely used in practice. In fact, when solving the 2D incompressible NS equations, we often use the vorticity-streamfunction formulation:
\begin{equation}
	\begin{alignedat}{1}
		\frac{\partial \Omega}{\partial t} + v_{i} \Omega_{,i} & = \frac{1}{Re} \nabla^{2} \Omega \\
		\nabla^{2} \psi & = \Omega
	\end{alignedat},
	\label{eq:vortex_stress}
\end{equation}
where \( \Omega \) and \( \psi \) are the vorticity and streamfunction, both of which are related to the velocity field:
\begin{equation}
\begin{aligned}
	\Omega & = u_{y} - v_{x} \\
	\frac{\partial \psi}{\partial x} & = -v \\
	\frac{\partial \psi}{\partial y} & = u
\end{aligned}.
\label{eq:vortex_velocity}
\end{equation}

It is easy to verify that the streamfunction automatically satisfies the incompressibility condition \( u_{i,i} = 0 \). To continue solving for the pressure, we need to solve the Poisson equation for the pressure:
\begin{equation}
\begin{aligned}
	\nabla^{2} p & = - \left[ \left( \frac{\partial u}{\partial x} \right)^{2} + 2 \frac{\partial v}{\partial x} \frac{\partial u}{\partial y} + \left( \frac{\partial v}{\partial y} \right)^{2} \right] \\
	& = 2 \left[ \frac{\partial u}{\partial x} \frac{\partial v}{\partial y} - \frac{\partial u}{\partial y} \frac{\partial v}{\partial x} \right]
	\label{eq:poisson}.
\end{aligned}
\end{equation}
Note that if the pressure field is not of interest, solving \Cref{eq:poisson_equation} is not necessary for solving \Cref{eq:vortex_stress}.
By examining \Cref{eq:vortex_stress} and \Cref{eq:poisson}, we observe that the pressure \( p \) and the velocity field \( \boldsymbol{u} \) are decoupled.

\begin{figure}
	\begin{centering}
		\includegraphics[scale=0.37]{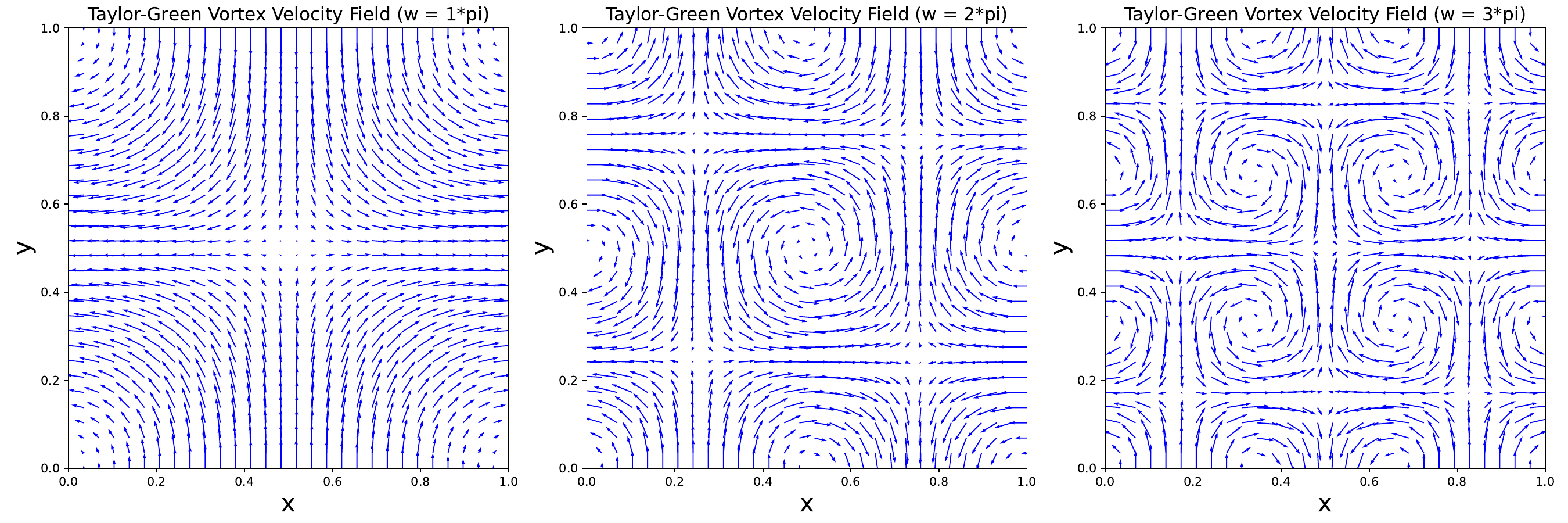}
		\par\end{centering}
	\caption{Introduction to the Taylor-Green vortex: the velocity vector field at different values of frequency $w$. The flow pattern of the Taylor-Green vortex typically involves multiple vortex structures, where the size and distribution of the vortices exhibit high symmetry. Over time, the vortices decay and eventually disappear.\label{fig:Taylor-green-vortex_intro} }
\end{figure}

We solve the Taylor-Green vortex problem shown in \Cref{fig:Taylor-green-vortex_intro}. The Taylor-Green vortex is a classic analytical solution to the incompressible Navier-Stokes equations, commonly used to verify the accuracy of computational fluid dynamics algorithms. The analytical solution is given by:
\begin{equation}
	\begin{aligned}
		\psi & = \frac{1}{w} \exp \left( - \frac{2w^{2}}{Re} t \right) \cos(wx) \cos(wy) \\
		\Omega & = u_{y} - v_{x} = -2w \exp \left( - \frac{2w^{2}}{Re} t \right) \cos(wx) \cos(wy) \\
		u & = \frac{\partial \psi}{\partial y} = - \exp \left( - \frac{2w^{2}}{Re} t \right) \cos(wx) \sin(wy) \\
		v & = - \frac{\partial \psi}{\partial x} = \exp \left( - \frac{2w^{2}}{Re} t \right) \sin(wx) \cos(wy) \\
		p & = - \frac{1}{2} \exp \left( - \frac{4w^{2}}{Re} t \right) \left[ \cos^{2}(wx) + \cos^{2}(wy) \right]
	\end{aligned},
	\label{eq:exact_solution_taylor}
\end{equation}
where \( w \) is a constant that controls the frequency of the Taylor-Green vortex. It is easy to verify that \Cref{eq:exact_solution_taylor} satisfies \Cref{eq:vortex_stress}.

The initial and boundary conditions for the Taylor-Green vortex are set according to the analytical solution and are detailed in \ref{sec:Taylor-Green-Vortex_boundary_initial}.

The loss function for solving \Cref{eq:vortex_stress,eq:vortex_velocity,eq:poisson} using the strong form of PINNs is:
\begin{equation}
	\begin{aligned}
		\mathcal{L}_{pinns} & = \frac{\lambda_{p}}{N_{p}} \sum_{i=1}^{N_{p}} \left[ \left| \frac{\partial \Omega(\boldsymbol{x}_{i})}{\partial t} + u \Omega(\boldsymbol{x}_{i})_{,x} + v \Omega(\boldsymbol{x}_{i})_{,y} - \frac{1}{Re} \nabla^{2} \Omega(\boldsymbol{x}_{i}) \right|^{2} + \left| \nabla^{2} \psi(\boldsymbol{x}_{i}) - \Omega(\boldsymbol{x}_{i}) \right|^{2} \right] \\
		& + \frac{\lambda_{\psi}^{b}}{N_{\psi}^{b}} \sum_{i=1}^{N_{\psi}^{b}} \left| \psi(\boldsymbol{x}_{i}) - \bar{\psi}(\boldsymbol{x}_{i}) \right|^{2} + \frac{\lambda_{\Omega}^{b}}{N_{\Omega}^{b}} \sum_{i=1}^{N_{\Omega}^{b}} \left| \Omega(\boldsymbol{x}_{i}) - \bar{\Omega}(\boldsymbol{x}_{i}) \right|^{2} \\
		& + \frac{\lambda_{u}^{b}}{N_{u}^{b}} \sum_{i=1}^{N_{u}^{b}} \left| u(\boldsymbol{x}_{i}) - \bar{u}(\boldsymbol{x}_{i}) \right|^{2} + \frac{\lambda_{v}^{b}}{N_{v}^{b}} \sum_{i=1}^{N_{v}^{b}} \left| v(\boldsymbol{x}_{i}) - \bar{v}(\boldsymbol{x}_{i}) \right|^{2} \\
		& + \frac{\lambda_{\psi}^{i}}{N_{\psi}^{i}} \sum_{i=1}^{N_{\psi}^{i}} \left| \psi(\boldsymbol{x}_{i}) - \bar{\psi}(\boldsymbol{x}_{i}) \right|^{2} + \frac{\lambda_{\Omega}^{i}}{N_{\Omega}^{i}} \sum_{i=1}^{N_{\Omega}^{i}} \left| \Omega(\boldsymbol{x}_{i}) - \bar{\Omega}(\boldsymbol{x}_{i}) \right|^{2} \\
		& + \frac{\lambda_{u}^{i}}{N_{u}^{i}} \sum_{i=1}^{N_{u}^{i}} \left| u(\boldsymbol{x}_{i}) - \bar{u}(\boldsymbol{x}_{i}) \right|^{2} + \frac{\lambda_{v}^{i}}{N_{v}^{i}} \sum_{i=1}^{N_{v}^{i}} \left| v(\boldsymbol{x}_{i}) - \bar{v}(\boldsymbol{x}_{i}) \right|^{2},
	\end{aligned}\label{eq:loss_function_vortex_stream}
\end{equation}
where \( N_{p} \) is the total number of collocation points for the PDEs loss in the domain; \( N_{\psi}^{b} \), \( N_{\Omega}^{b} \), \( N_{u}^{b} \), and \( N_{v}^{b} \) are the total numbers of collocation points for the boundary conditions of \( \psi \), \( \Omega \), \( u \), and \( v \), respectively; and \( N_{\psi}^{i} \), \( N_{\Omega}^{i} \), \( N_{u}^{i} \), and \( N_{v}^{i} \) are the total numbers of collocation points for the initial conditions of \( \psi \), \( \Omega \), \( u \), and \( v \), respectively. All collocation points are uniformly and randomly distributed. We set \( N_{p} = 1000 \), \( N_{\psi}^{b} = 100 \), \( N_{\Omega}^{b} = 100 \), \( N_{u}^{b} = 100 \), and \( N_{v}^{b} = 100 \), while \( N_{\psi}^{i} = 100 \), \( N_{\Omega}^{i} = 100 \), \( N_{u}^{i} = 100 \), and \( N_{v}^{i} = 100 \). 

The weights \( \lambda_{p} \), \( \lambda_{\psi}^{b} \), \( \lambda_{\Omega}^{b} \), \( \lambda_{u}^{b} \), and \( \lambda_{v}^{b} \) correspond to the PDEs loss in the domain and the boundary condition losses for \( \psi \), \( \Omega \), \( u \), and \( v \), respectively. Similarly, \( \lambda_{\psi}^{i} \), \( \lambda_{\Omega}^{i} \), \( \lambda_{u}^{i} \), and \( \lambda_{v}^{i} \) are the weights for the initial condition losses of \( \psi \), \( \Omega \), \( u \), and \( v \), respectively. Although there are techniques for adjusting hyperparameters \citep{ill_gradient,NTK_PINN,NTK_to_get_hyperparameter_of_PINN}, determining the optimal hyperparameters remains challenging. Therefore, we manually set all hyperparameters \( \lambda_{p} \), \( \lambda_{\psi}^{b} \), \( \lambda_{\Omega}^{b} \), \( \lambda_{u}^{b} \), \( \lambda_{v}^{b} \), \( \lambda_{\psi}^{i} \), \( \lambda_{\Omega}^{i} \), \( \lambda_{u}^{i} \), and \( \lambda_{v}^{i} \) to 1.

The MLP architecture is \( [3, 100, 100, 100, 100, 2] \), where the inputs are the coordinates \( x \), \( y \), and \( t \), and the outputs are the stream function \( \psi \) and the vorticity \( \Omega \). The hidden layers of the neural network use the \(\tanh\) activation function, while the final layer has no activation function. For objective comparison, we use the Adam optimizer with a learning rate of 0.001. 

In the loss function \Cref{eq:loss_function_vortex_stream}, note that the velocities \( u \) and \( v \) are derived from the stream function using \Cref{eq:vortex_velocity} to transform the stream function into the velocity field.

\begin{figure}
	\begin{centering}
		\includegraphics[scale = 0.42]{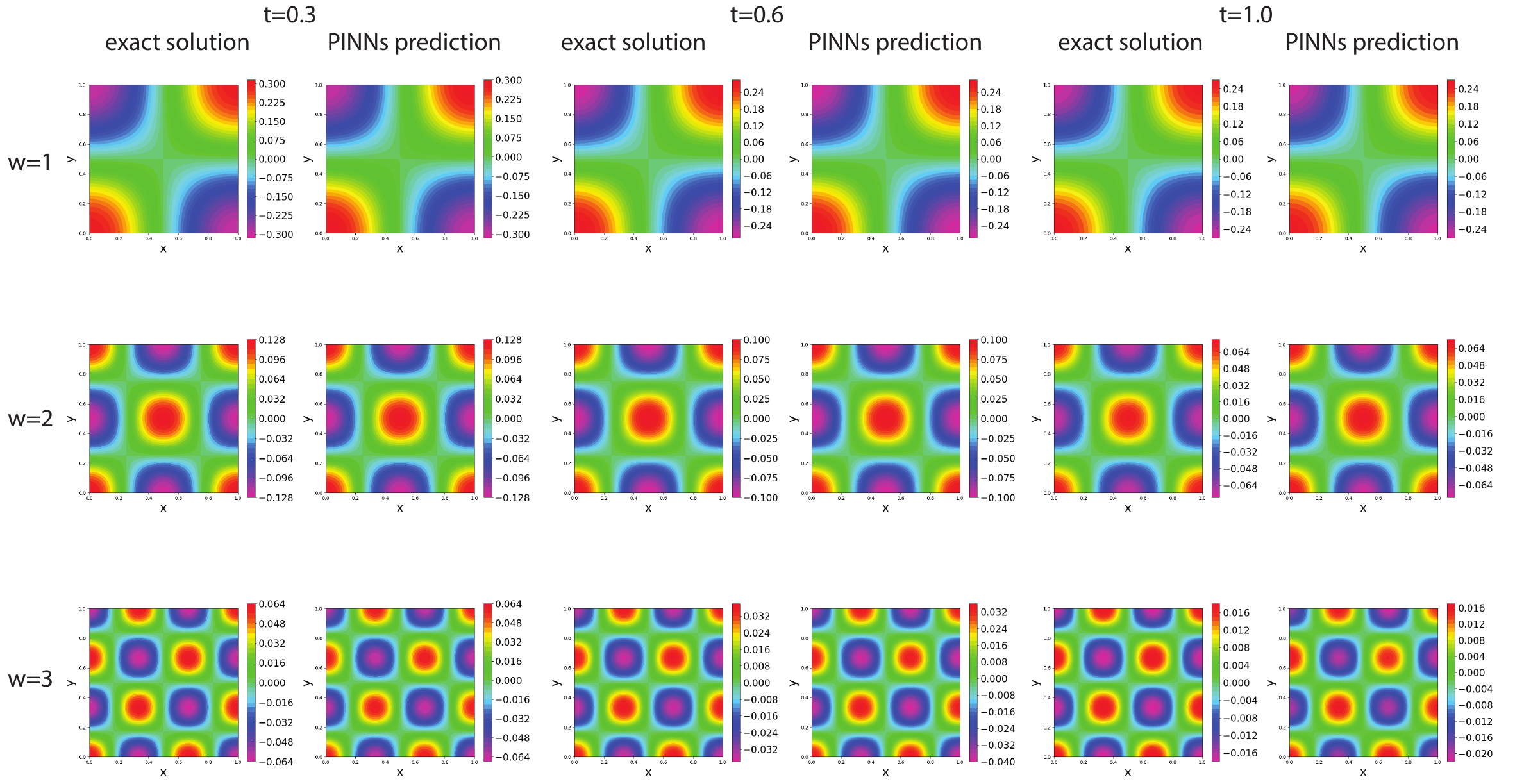}
		\par\end{centering}
	\caption{Performance of PINNs in the strong form on the stream function: for different $w$ values ($w=1.0*\pi$, $w=2.0*\pi$, and $w=3.0*\pi$), at different times ($t=0.3$, $t=0.6$, and $t=1.0$)\label{fig:PINNs_stream_function}}
\end{figure}

\begin{figure}
	\begin{centering}
		\includegraphics[scale = 0.42]{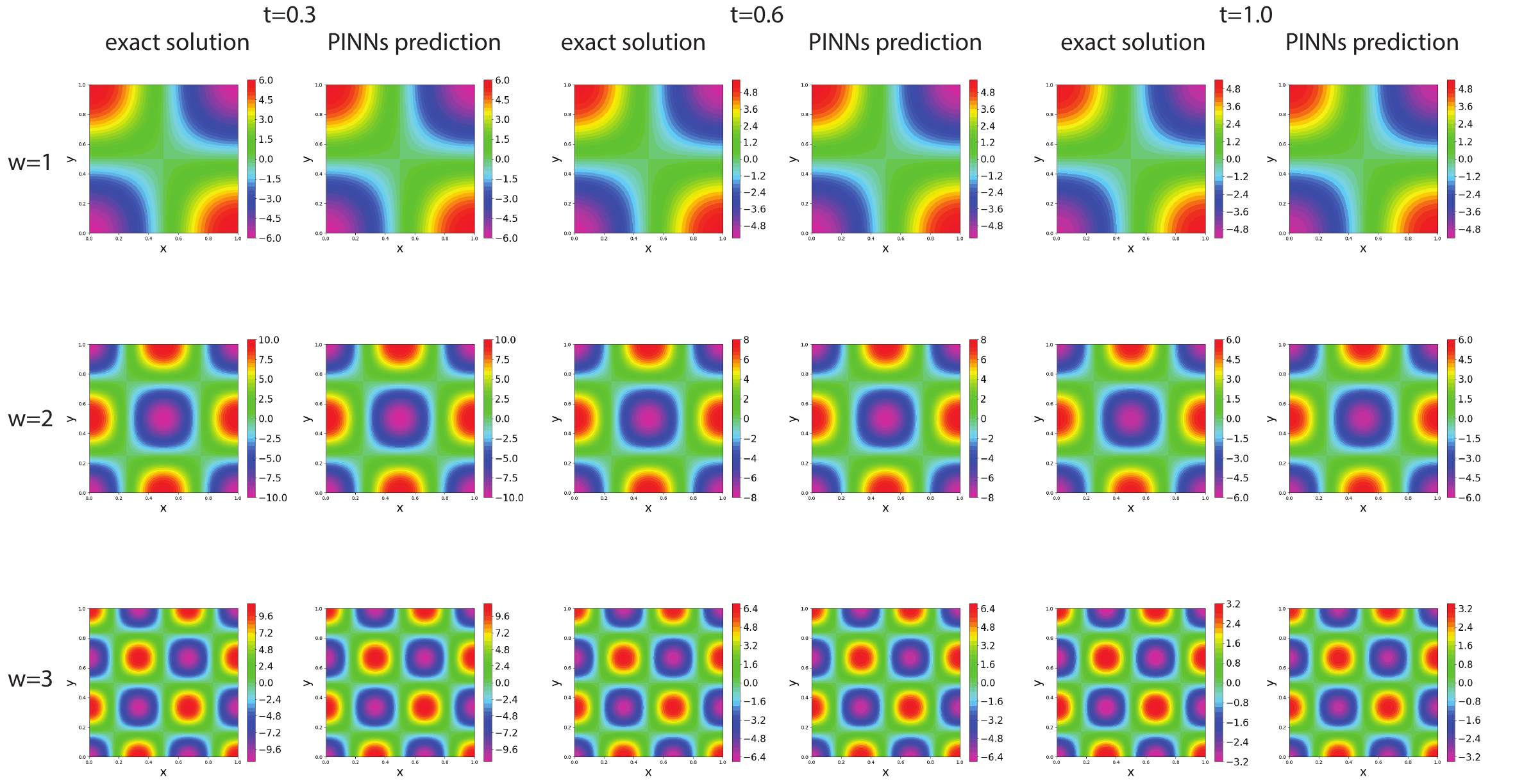}
		\par\end{centering}
	\caption{Performance of PINNs in the strong form on vorticity: for different $w$ values ($w=1.0*\pi$, $w=2.0*\pi$, and $w=3.0*\pi$), at different times ($t=0.3$, $t=0.6$, and $t=1.0$)\label{fig:PINNs_vorticity}}
\end{figure}

\Cref{fig:PINNs_stream_function} shows a comparison of the exact solution and the predicted stream function from PINNs at different times (0.3, 0.6, 1.0) and different $w$ values. \Cref{fig:PINNs_vorticity} demonstrates a comparison of the exact solution and the predicted vorticity from PINNs at different times (0.3, 0.6, 1.0). We can observe that PINNs can simulate the Taylor Green vortex accurately.

\begin{figure}
	\begin{centering}
		\includegraphics[scale = 0.45]{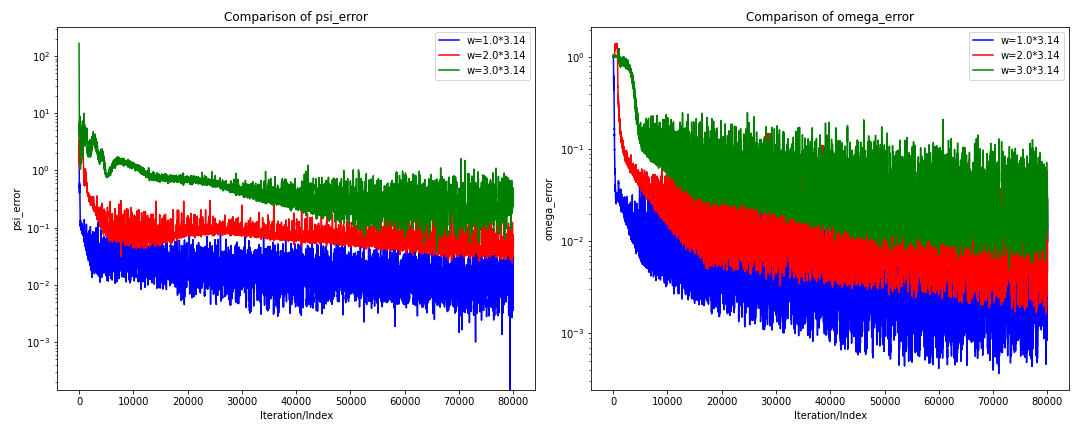}
		\par\end{centering}
	\caption{Evolution of the relative error $\mathcal{L}_{2}$ of PINNs in the strong form on the stream function and vorticity\label{fig:PINNs_taylor_error}}
\end{figure}

To more accurately demonstrate the performance of PINNs on the Taylor Green vortex, we need to observe the evolution of the relative error $\mathcal{L}_{2}$ as the iterations progress. The calculation formula for $\mathcal{L}_{2}$ is as follows:
\begin{equation}
\mathcal{L}_{2} = \frac{||\boldsymbol{u}_{exact}-\boldsymbol{u}_{pred}||_{2}}{||\boldsymbol{u}_{exact}||_{2}}
\end{equation}
where $||\cdot||_{2}$ is the $L_2$ norm. \Cref{fig:PINNs_taylor_error} shows the evolution of the relative error of PINNs as the iterations progress. We can clearly see that as $w$ increases, the accuracy of PINNs decreases. This is due to the spectral bias of fully connected neural networks \citet{NTK_PINN}. Spectral bias refers to the phenomenon where fully connected neural networks converge more strongly on low frequencies than on high frequencies. Moreover, the physical meaning of $w$ is the frequency of the solution space, meaning that as $w$ increases, the frequency of the target solution becomes higher. This means that fully connected neural networks perform less well when handling target solutions with higher frequencies.

\begin{figure}
	\begin{centering}
		\includegraphics[scale=0.32]{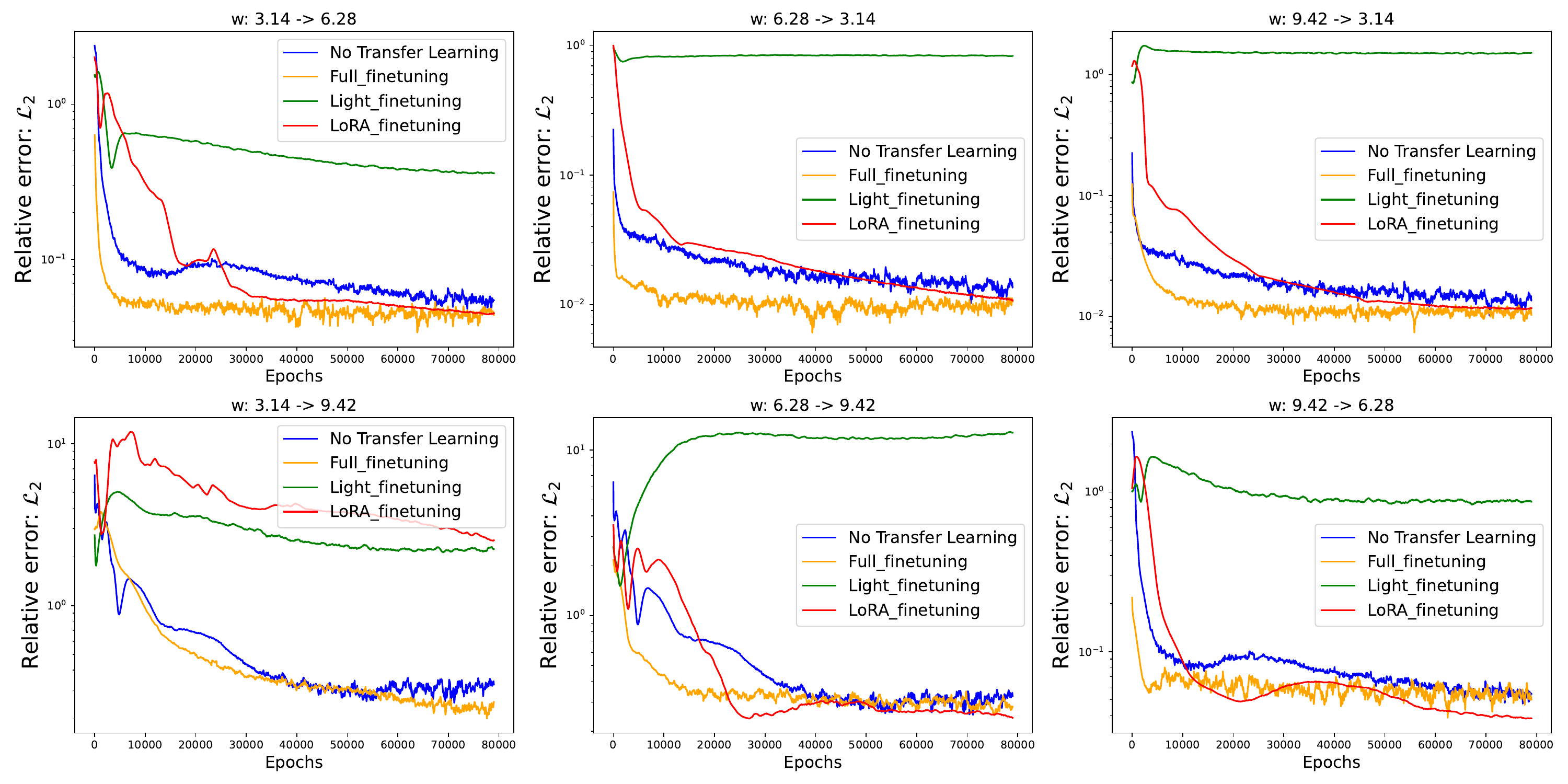}
		\par\end{centering}
	\caption{Evolution of the relative error $\mathcal{L}_{2}$ of transfer learning on the stream function in PINNs' strong form: $w = X \to Y$ indicates that in PINNs, pre-training is performed on the source domain $w = X$, and then transfer learning is applied to the target domain $w = Y$.
		\label{fig:transfer_PINNs_stream_L2}}
\end{figure}

\begin{figure}
	\begin{centering}
		\includegraphics[scale=0.32]{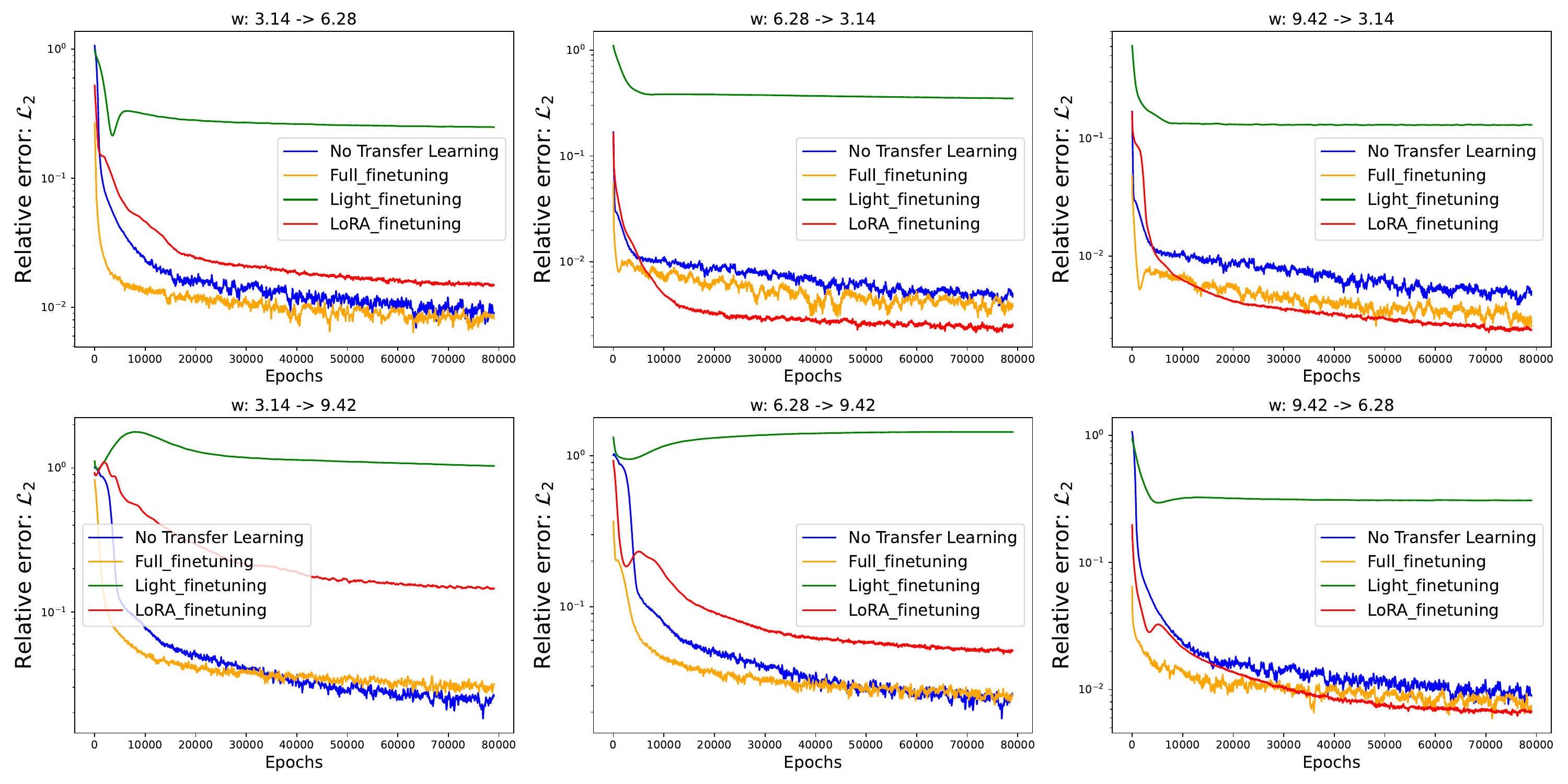}
		\par\end{centering}
	\caption{Evolution of the relative error $\mathcal{L}_{2}$ of transfer learning on the vorticity in PINNs' strong form: $w = X \to Y$ indicates that in PINNs, pre-training is performed on the source domain $w = X$, and then transfer learning is applied to the target domain $w = Y$.
		\label{fig:transfer_PINNs_vortex_L2}}
\end{figure}

Next, we are naturally curious about the generalization performance of transfer learning on the boundary conditions in PINNs' strong form. For boundary condition generalization, we adopt a transfer learning scheme where $w$ is randomly selected from $\{\pi, 2\pi, 3\pi\}$ as the source domain, and the parameters are transferred to the other two $w$ values as the target domain. For example, PINNs are trained on $w=\pi$, and then the parameters are directly transferred to $w=2\pi$ or $w=3\pi$ as the initial values. Here, we follow the transfer learning schemes described in \Cref{sec:Method:-Transfer-learning}, which are full finetuning, lightweight finetuning, and LoRA. \Cref{fig:transfer_PINNs_stream_L2} and \Cref{fig:transfer_PINNs_vortex_L2} show the evolution trends of the relative error of different transfer learning methods on different boundary conditions. No transfer learning refers to the evolution trend of the relative error of PINNs without transfer learning. Full finetuning refers to fully fine-tuning the pre-trained parameters. Full finetuning refers to fully fine-tuning the pre-trained parameters. 
We adopt the method of lightweight finetuning as described in \citet{goswami2020transfer, chakraborty2021transfer, xu2023transfer}, which only trains the final layer of the neural network. However, we found that the lightweight approach performed poorly. This is because the neural network of PINNs does not have the hierarchical features typical of CNNs used in computer vision \citep{zeiler2014visualizing}.
 LoRA trains only the hidden layers in the middle, without training the input and output layer, i.e., LoRA only trains the layers $[3,100,100,100,100,2]$ on $[100,100,100,100]$, and we set the rank of LoRA to 4. 
\Cref{subsec:The-rank-in-LoRA} shows the effects of different ranks in LoRA.
 
It is worth noting that the full finetuning and lightweight finetuning are the same when epoch is 0 in transfer learning. However, due to the use of curve averaging smoothing technique, the starting points for full finetuning and lightweight finetuning in \Cref{fig:transfer_PINNs_stream_L2} and \Cref{fig:transfer_PINNs_vortex_L2} may differ.
In addition, the initial value of LoRA in transfer learning is different from full finetuning and light
finetuning because LoRA involves additional initialization parameters $\boldsymbol{A}$ and $\boldsymbol{B}$ shown in \Cref{eq:lora_computation}.

\Cref{fig:transfer_PINNs_stream_L2} and \Cref{fig:transfer_PINNs_vortex_L2} show that lightweight finetuning performs the worst, because the change in the problem is too large, but the number of parameters the neural network can learn is too small, which is insufficient to meet the actual needs. Additionally, we observe an interesting phenomenon: when the pre-trained $w$ is larger than the target $w$, LoRA generally performs better than when the pre-trained $w$ is smaller than the target $w$ (the pre-trained $w$ refers to the $w$ on which PINNs are initially trained, and the target $w$ refers to the $w$ the parameters are transferred to). This may be because the transfer between low and high frequencies in fully connected neural networks is not bidirectional; that is, high-to-low frequency transfer is more effective than low-to-high frequency transfer. This could also be caused by spectral bias: when the pre-trained $w$ is low-frequency, the neural network has already converged to a stable loss value, and if we transfer it to a high-frequency problem, the well-converged network may not easily escape local minima. However, when pre-trained on high frequencies, the neural network has not converged to a stable local minimum due to the network's poorer fitting ability for high frequencies. In this case, the network can better transfer to lower frequencies. In summary, full finetuning performs well in all scenarios and outperforms the case with no transfer learning.

Although full finetuning generally performs the best, the number of learnable parameters in full finetuning is larger than in lightweight finetuning and LoRA, which makes each epoch take more time. \Cref{tab:PINNs_strong_transfer} shows the efficiency and the number of learnable parameters for different transfer learning methods. All computations are carried out on a single Nvidia RTX 4060 Ti GPU with 16GB memory. We noticed an odd phenomenon: the efficiency of lightweight finetuning does not show a significant improvement over full finetuning. This is because when PINNs approximate the differential operators using Automatic Differentiation \citet{automatic_differential} during backpropagation, a new computation graph is constructed, which reduces the gains of lightweight finetuning. We explain this in \ref{sec:Computation-graph-of_PINN}. Moreover, LoRA turns out to be less efficient than full finetuning. Although LoRA has fewer learnable parameters than full finetuning, the internal matrix computations in LoRA are actually more than in full finetuning, which we explain in \ref{sec:Computation-of-LoRA}.
We tested the rank in LoRA to two extreme cases, $r = 1$ and $r = 100$. The results in \Cref{tab:PINNs_strong_transfer} show that when $r = 100$, the $\mathcal{L}_{2}$ error is close to that of full finetuning, as $r = 100$ is equal to full finetuning.

\begin{table}
	\caption{Accuracy and efficiency of different transfer learning methods: The relative error $\mathcal{L}_{2}$ is recorded at the 30,000th iteration. $w=X \rightarrow Y$ indicates that PINNs are pre-trained on $w=X$ as the source domain, then transfer learning is performed on $w=Y$ as the target domain.  The bold numbers represent the best solutions among all transfer learning solutions. \label{tab:PINNs_strong_transfer}}
		\begin{adjustbox}{max width=\textwidth}
	\centering
	\begin{tabular}{ccccccc}
		\toprule
		& No transfer & Full\_finetuning & Lightweight\_finetuning (last) & LoRA ($r=1$) & LoRA ($r=4$) & LoRA ($r=100$) \\
		\midrule
		Time (s, 1000 Epochs) & 14.6 & 14.6 & \textbf{12.3} & 14.6 & 14.8 & 14.9 \\
		Trainable parameters & 30902 & 30902 & \textbf{202} & 900 & 2700 & 60300 \\
		Vorticity: $\mathcal{L}_{2}$ ($w=\pi \rightarrow 2\pi$) & 0.014540 & \textbf{0.010842} & 0.27463 & 0.10166 & 0.021114 & 0.014770 \\
		Vorticity: $\mathcal{L}_{2}$ ($w=\pi \rightarrow 3\pi$) & 0.039597 & \textbf{0.037837} & 1.1711 & 1.2465 & 0.21432 & 0.079696 \\
		Vorticity: $\mathcal{L}_{2}$ ($w=2\pi \rightarrow \pi$) & 0.0078504 & 0.0051498 & 0.38377 & 0.0096263 & 0.0029064 & \textbf{0.0039211} \\
		Vorticity: $\mathcal{L}_{2}$ ($w=2\pi \rightarrow 3\pi$) & 0.039597 & 0.033586 & 1.51422 & 0.28875 & 0.070175 & \textbf{0.0040570} \\
		Vorticity: $\mathcal{L}_{2}$ ($w=3\pi \rightarrow \pi$) & 0.0078504 & 0.0049677 & 0.21606 & 0.017968 & 0.0035354 & \textbf{0.0049142} \\
		Vorticity: $\mathcal{L}_{2}$ ($w=3\pi \rightarrow 2\pi$) & 0.014540 & 0.0098385 & 0.21788 & 0.032850 & 0.010454 & \textbf{0.0065526} \\
		Stream function: $\mathcal{L}_{2}$ ($w=\pi \rightarrow 2\pi$) & 0.088086 & \textbf{0.047082} & 0.41158 & 0.60451 & 0.060649 & 0.10954 \\
		Stream function: $\mathcal{L}_{2}$ ($w=\pi \rightarrow 3\pi$) & 0.43473 & \textbf{0.36305} & 3.2088 & 15.529 & 4.0337 & 3.0178 \\
		Stream function: $\mathcal{L}_{2}$ ($w=2\pi \rightarrow \pi$) & 0.018961 & \textbf{0.010571} & 0.65634 & 0.021319 & 0.023206 & 0.01343 \\
		Stream function: $\mathcal{L}_{2}$ ($w=2\pi \rightarrow 3\pi$) & 0.43473 & 0.33061 & 3.5122 & 1.5075 & \textbf{0.25494} & 0.42985 \\
		Stream function: $\mathcal{L}_{2}$ ($w=3\pi \rightarrow \pi$) & 0.018961 & 0.012136 & 0.64111 & 0.097851 & 0.019430 & \textbf{0.011005} \\
		Stream function: $\mathcal{L}_{2}$ ($w=3\pi \rightarrow 2\pi$) & 0.088086 & 0.057240 & 0.58412 & 0.62227 & 0.060155 & \textbf{0.021474} \\
		\bottomrule
	\end{tabular}
\end{adjustbox}
\end{table}

\subsection{Transfer learning for different materials\label{subsec:Transfer-learning-material}}

\begin{figure}
	\begin{centering}
		\includegraphics[scale=0.55]{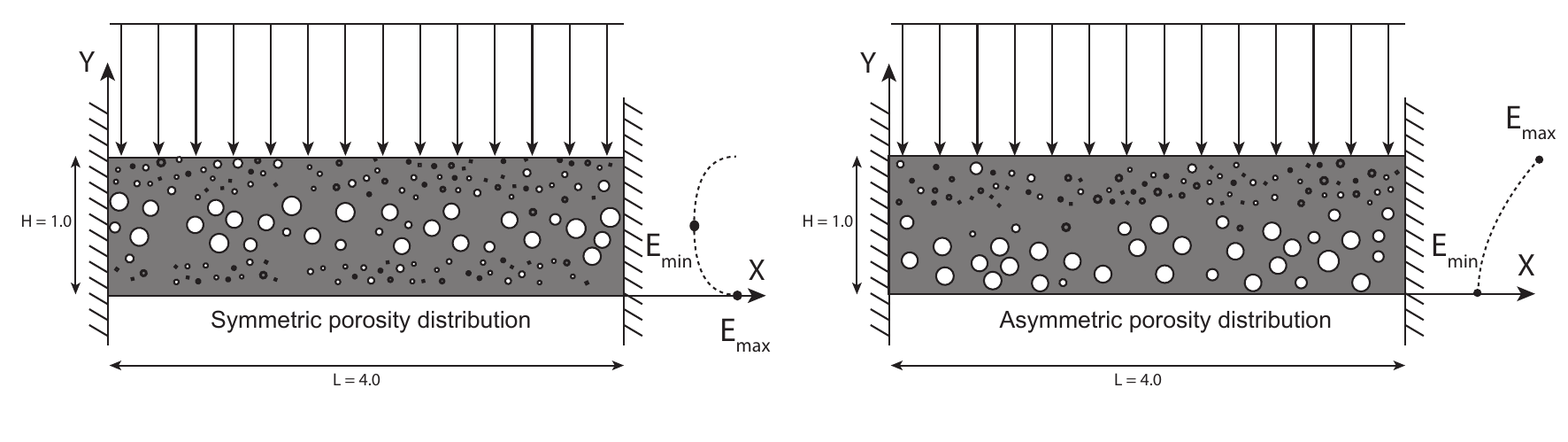}
		\par\end{centering}
	\caption{Schematic of functionally graded porous beams: Both ends are clamped. The upper surface of the beam is subjected to a uniform load $f=1N/m$. The symmetric and asymmetric porosity distributions are two different material distributions for functionally graded porous beams.\label{fig:Functionally-graded-porous}}
\end{figure}

In this subsection, we verify the generalization capability of transfer learning on different materials using functionally graded porous beams as a numerical example. The characteristic of functionally graded porous beams is that the material density is an inhomogeneous field, as shown in \Cref{fig:Functionally-graded-porous}. The elastic modulus and shear modulus of the functionally graded porous beams are proportional to the density.

The elastic modulus and shear modulus for the symmetric porosity distribution are given by:
\begin{equation}
	\begin{aligned}
		E(x,y) & = E_{max} - \cos\left[\pi \left(\frac{y}{H} - \frac{1}{2}\right)\right] (E_{max} - E_{min}) \\
		G(x,y) & = G_{max} - \cos\left[\pi \left(\frac{y}{H} - \frac{1}{2}\right)\right] (G_{max} - G_{min})
	\end{aligned}.
	\label{eq:sym_porosity_distribution}
\end{equation}

For the asymmetric porosity distribution, the elastic modulus and shear modulus are:
\begin{equation}
	\begin{aligned}
		E(x,y) & = E_{max} - \cos\left(\pi \frac{y}{2H}\right) (E_{max} - E_{min}) \\
		G(x,y) & = G_{max} - \cos\left(\pi \frac{y}{2H}\right) (G_{max} - G_{min})
	\end{aligned},
	\label{eq:asym_porosity_distribution}
\end{equation}
where $E_{max}$ and $E_{min}$ represent the maximum and minimum elastic moduli, respectively, and $G_{max}$ and $G_{min}$ represent the maximum and minimum shear moduli. Note that the material density field only depends on $y$ and not on $x$, which means the mass density varies only in the $y$ direction. The symmetric porosity distribution has the highest density at the ends ($y=0$ and $y=H$) and the lowest at the center ($y=H/2$), while the asymmetric porosity distribution has the highest density at the top and the lowest at the bottom.

The functionally graded porous beams are governed by the classical elastic mechanics PDEs, with the governing equations given by:
\begin{equation}
	\begin{aligned}
		\sigma_{\alpha\beta,\beta} + f_{\alpha} & = 0 \\
		\sigma_{\alpha\beta} & = 2G \varepsilon_{\alpha\beta} + \lambda \varepsilon_{kk} \delta_{\alpha\beta} \\
		\varepsilon_{\alpha\beta} & = \frac{1}{2} \left( u_{\alpha,\beta} + u_{\beta,\alpha} \right)
	\end{aligned},
	\label{eq:strong_form_elasticity}
\end{equation}
where $\alpha, \beta = x, y$. We solve the plane stress problem, thus $\varepsilon_{zz} = -\frac{\upsilon}{1-\upsilon} \left( \varepsilon_{xx} + \varepsilon_{yy} \right)$. The Lamé coefficients are given by:
\begin{equation}
	\begin{aligned}
		\lambda & = \frac{\upsilon E}{(1 + \upsilon)(1 - 2\upsilon)} \\
		G & = \frac{E}{2(1 + \upsilon)}
	\end{aligned},
\end{equation}
where $E$ is the elastic modulus and $\upsilon$ is Poisson's ratio.

\Cref{eq:strong_form_elasticity} is the strong form of the PDEs. Solving this with the strong form in PINNs requires tuning many hyperparameters, and the highest derivative order is second \cite{wang2022cenn}. Since we can use the variational principle to convert \Cref{eq:strong_form_elasticity} into its corresponding energy form, solving with the PINNs energy form (Deep Energy Method) is more suitable because it significantly reduces the hyperparameters and lowers the highest derivative order to 1. The energy form of  \Cref{eq:strong_form_elasticity} is:
\begin{equation}
	\mathcal{L} = \int_{\Omega} \frac{1}{2} \sigma_{\alpha\beta} \varepsilon_{\alpha\beta} \, d\Omega - \int_{\Omega} f_{\alpha} u_{\alpha} \, d\Omega - \int_{\Gamma^{\boldsymbol{t}}} \bar{t}_{\alpha} u_{\alpha} \, d\Gamma
	\label{eq:DEM_elasticity},
\end{equation}
where $\bar{t}_{\alpha}$ is the force boundary condition on the Neumann boundary $\Gamma^{\boldsymbol{t}}$. It is easy to verify that $\delta \mathcal{L} = 0$ is equivalent to \Cref{eq:strong_form_elasticity}.

In the DEM, we set $\mathcal{L}$ in \Cref{eq:DEM_elasticity} as the loss function. It is worth noting that the displacement field in the DEM must satisfy the essential boundary conditions in advance. Considering the essential boundary conditions shown in \Cref{fig:Functionally-graded-porous}, which are clamped, the displacement field is assumed to be:
\begin{equation}
	u(x,y) = NN(x,y;\boldsymbol{\theta}) * x * (L - x)
	\label{eq:dis_admiss_elasticity},
\end{equation}
where $NN(x, y; \boldsymbol{\theta})$ is an MLP neural network, and $\boldsymbol{\theta}$ represents the learnable parameters of the network. The structure of the MLP is $[2, 100, 100, 100, 100, 2]$, with $\tanh$ as the activation function in the hidden layers, and no activation function in the final layer.
It is easy to verify that \Cref{eq:dis_admiss_elasticity} satisfies the clamped boundary conditions at both ends ($x=0$ and $x=L$).

\begin{figure}
	\begin{centering}
		\includegraphics[scale=0.5]{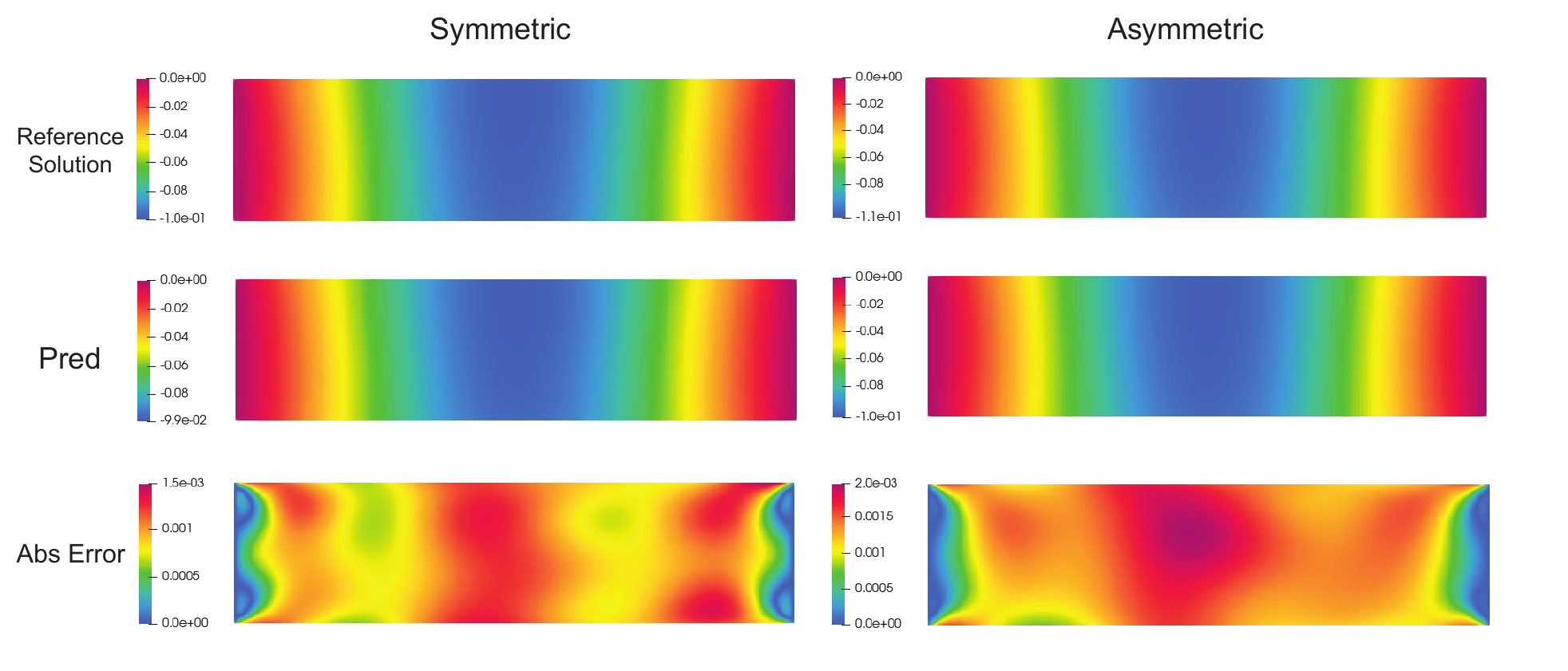}
		\par\end{centering}
	\caption{Prediction contour maps of the displacement field $u_{y}$ for symmetric (first column) and asymmetric (second column) porosity distributions using the PINNs energy form: The first row shows the reference solution obtained from isogeometric analysis; the second row shows the DEM prediction; the third row shows the absolute error.\label{fig:DEM_material_contourf}}
\end{figure}

In \Cref{eq:sym_porosity_distribution} and \Cref{eq:asym_porosity_distribution}, $E_{max}$ and $E_{min}$ are set to 200 and 100, respectively. 
Poisson's ratio is $1/3$.
We set the optimizer (Adam) for objective comparison with a learning rate of 0.001. Given that, under this load, the $y$-direction displacement $u_{y}$ is the main displacement field, the predicted contour map of the DEM displacement field $u_{y}$ is shown in \Cref{fig:DEM_material_contourf}. We observe that the DEM-predicted displacement field aligns well with the exact solution. Note that we use isogeometric analysis (IGA) as the reference solution \cite{eshaghi2025applications}. The collocation points for both DEM and IGA are arranged on a uniform grid with a spacing of 0.01, resulting in a $401 \times 101$ grid.
The integration scheme in DEM is the trapezoidal integration method, rather than the traditional Monte Carlo integration, because the integration scheme significantly affects the accuracy of DEM \citet{wang2025physics}. Monte Carlo integration does not perform as well as methods like trapezoidal, Simpson's numerical integrations, or triangular integration in DEM \citet{PINN_hyperelasticity}.

\begin{figure}
	\begin{centering}
		\includegraphics[scale=0.45]{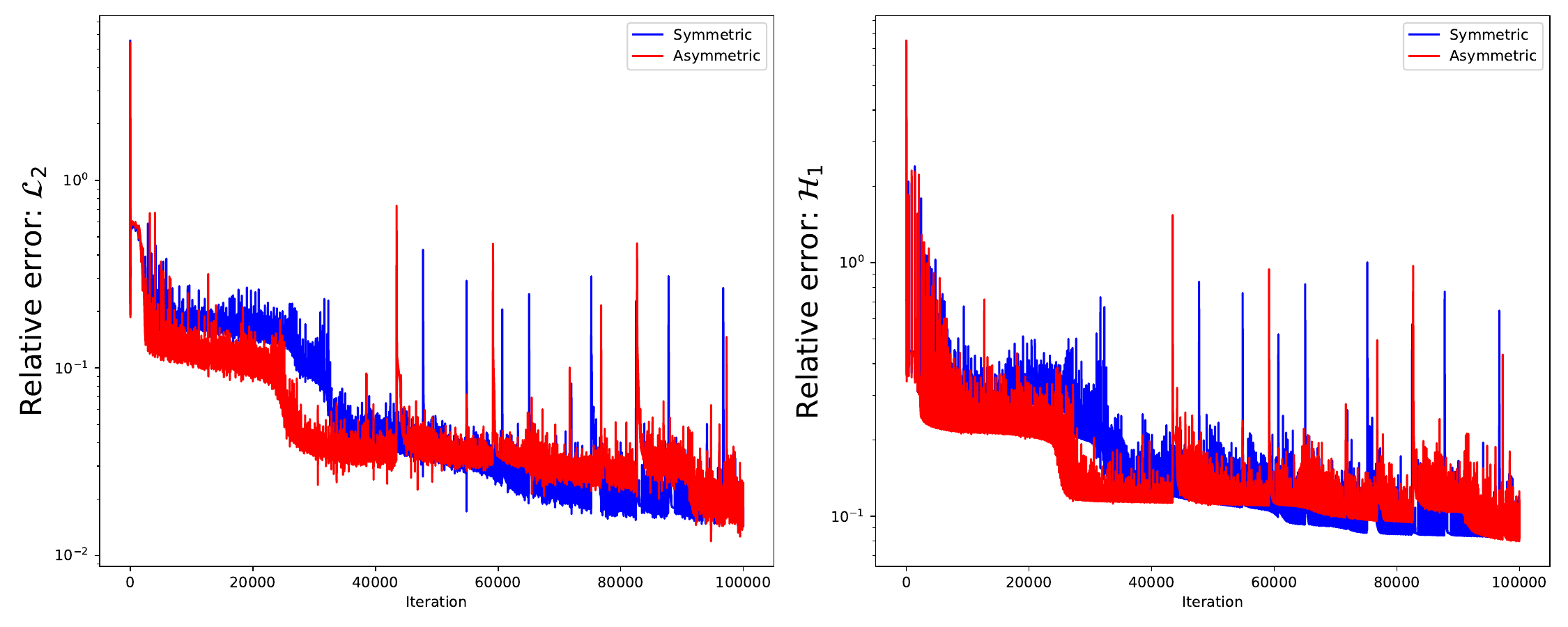}
		\par\end{centering}
	\caption{Comparison of relative displacement $u_{y}$ error $\mathcal{L}_{2}$ (left) and Von Mises stress error $\mathcal{H}_{1}$ (right) for symmetric and asymmetric porosity distributions using the PINNs energy form.\label{fig:sym_asym_L2_H1_error}}
\end{figure}

\Cref{fig:sym_asym_L2_H1_error} shows the relative errors of the DEM displacement $\mathcal{L}_{2}$ and Von Mises stress $\mathcal{H}_{1}$ as they evolve with iterations. The Von Mises stress is calculated as:
\begin{equation}
Mises = \sqrt{\frac{3}{2} (\sigma_{ij} - \frac{1}{3} \sigma_{kk} \delta_{ij})(\sigma_{ij} - \frac{1}{3} \sigma_{mm} \delta_{ij})}\label{eq:von_mises}.
\end{equation}
We observe that the relative error of Von Mises stress $\mathcal{H}_{1}$ is larger than the displacement error $\mathcal{L}_{2}$ due to the infinite stress at corner points. Theoretically, stress solutions at corner points will always have errors. Additionally, since stress requires the first derivative of the displacement field, the stress error will naturally be higher than the displacement error (the higher the derivative order in PINNs, the larger the error \cite{wang2023dcm}).

\begin{figure}
	\begin{centering}
		\includegraphics[scale=0.33]{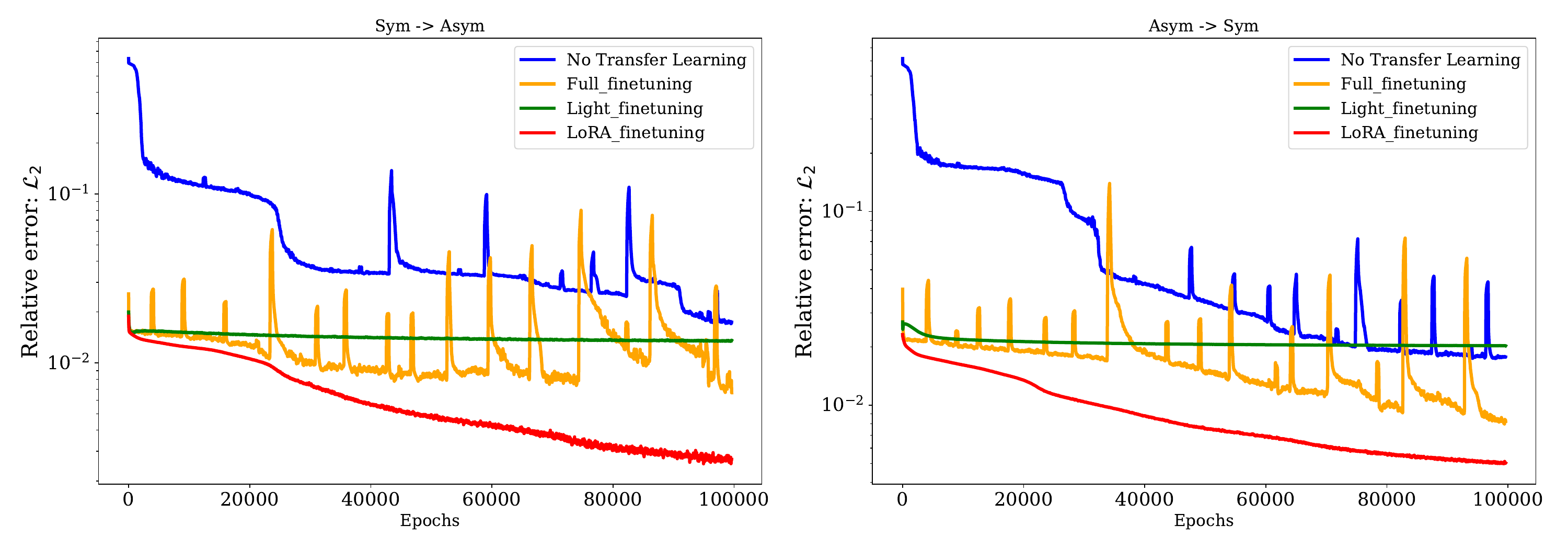}
		\par\end{centering}
	\caption{Relative error evolution of $\mathcal{L}_{2}$ ($u_{y}$) for material generalization in the PINNs energy form: Sym and Asym represent the symmetric and asymmetric porosity distributions, respectively. $Sym \rightarrow Asym$ indicates that the symmetric porosity distribution is the source domain, and the asymmetric porosity distribution is the target domain. Lightweight finetuning only trains the last layer of the neural network, while LoRA finetuning uses a rank of 4.\label{fig:transfer_sym_asym_L2}}
\end{figure}

\begin{figure}
	\begin{centering}
		\includegraphics[scale=0.33]{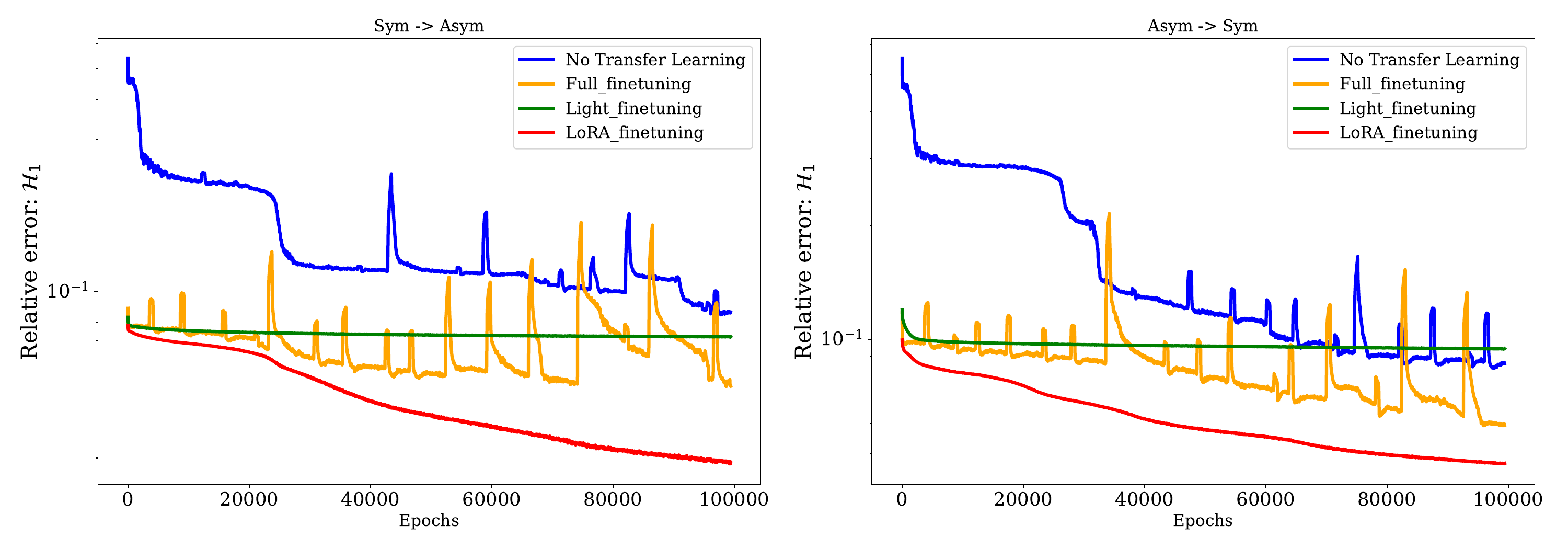}
		\par\end{centering}
	\caption{Relative error evolution of $\mathcal{H}_{1}$ (Von Mises) for material generalization in the PINNs energy form: Sym and Asym represent the symmetric and asymmetric porosity distributions, respectively. $Sym \rightarrow Asym$ indicates that the symmetric porosity distribution is the source domain, and the asymmetric porosity distribution is the target domain. Lightweight finetuning only trains the last layer of the neural network, while LoRA finetuning uses a rank of 4.\label{fig:transfer_sym_asym_H1}}
\end{figure}

Next, we demonstrate the transfer learning effects for different materials using the PINNs energy form. The idea behind our transfer learning approach is to use the symmetric porosity distribution as the source domain, i.e., the parameters of the DEM model after 100,000 epochs are used as the initial values for the neural network parameters. The asymmetric porosity distribution is then treated as the target domain, and vice versa.  \Cref{fig:transfer_sym_asym_L2} and \Cref{fig:transfer_sym_asym_H1} show the effects of different transfer learning methods on the generalization of the materials. In the case of lightweight fine-tuning, only the last layer of the neural network is trained, while the other parameters remain fixed. In the LoRA method, we set the rank to 4. We observe that among all the transfer learning approaches, LoRA yields the best performance.

To further compare the precision and efficiency of different transfer learning approaches for material generalization, we vary the rank in LoRA to two extreme values: $1$ and $100$. \Cref{tab:transfer_sym_asym} presents the results of various transfer learning strategies. We observe that the lowest errors for material generalization are achieved when the rank in LoRA is set to 4. The performance of full fine-tuning and LoRA with rank 100 is similar, as a higher rank in LoRA approaches the effect of full fine-tuning, as shown in \Cref{fig:Parameter-based-transfer-learn}c. Additionally, lightweight fine-tuning performs poorly, which is consistent with the results in \Cref{tab:PINNs_strong_transfer}, and in some cases, it even increases the relative errors $\mathcal{L}_{2}$ and $\mathcal{H}_{1}$ compared to no transfer learning.

\begin{table}
	\caption{Accuracy and efficiency of different transfer learning approaches for material generalization: The relative errors $\mathcal{L}_{2}$ ($u_{y}$) and $\mathcal{H}_{1}$ (Von Mises) are computed at 100,000 iterations. $Sym \rightarrow Asym$ indicates that the symmetric porosity distribution is the source domain, and the asymmetric porosity distribution is the target domain. The reverse is also true.\label{tab:transfer_sym_asym}}
	\begin{adjustbox}{max width=\textwidth}
	\centering{}%
	\begin{tabular}{ccccccc}
		\toprule 
		& No transfer & Full\_finetuning & Lightweight\_finetuning (last) & LoRA ($r=1$) & LoRA ($r=4$) & LoRA ($r=100$)\tabularnewline
		\midrule
		Time (s, 1000 Epoches) & 32.41 & 32.41 & \textbf{30.30} & 31.77 & 31.81 & 32.3\tabularnewline
		Trainable parameters & 30802 & 30802 & \textbf{202} & 900 & 2700 & 60300\tabularnewline
		Displacement $\mathcal{L}_{2}$ relative error: $Sym \rightarrow Asym$ & 0.017427 & 0.0066538 & 0.013616 & 0.0044375 & \textbf{0.0026861} & 0.0035537\tabularnewline
		Displacement $\mathcal{L}_{2}$ relative error: $Asym \rightarrow Sym$ & 0.017698 & 0.0082514 & 0.020277 & 0.0087151 & \textbf{0.0050296} & 0.0086635\tabularnewline
		Stress $\mathcal{H}_{1}$ relative error: $Sym \rightarrow Asym$ & 0.086220 & 0.050413 & 0.072096 & 0.039482 & \textbf{0.028944} & 0.03093\tabularnewline
		Stress $\mathcal{H}_{1}$ relative error: $Asym \rightarrow Sym$ & 0.086337 & 0.059366 & 0.094390 & 0.061203 & \textbf{0.047018} & 0.057451\tabularnewline
		\bottomrule
	\end{tabular}
\end{adjustbox}
\end{table}

\subsection{Transfer learning for different geometries}

\begin{figure}
	\begin{centering}
		\includegraphics{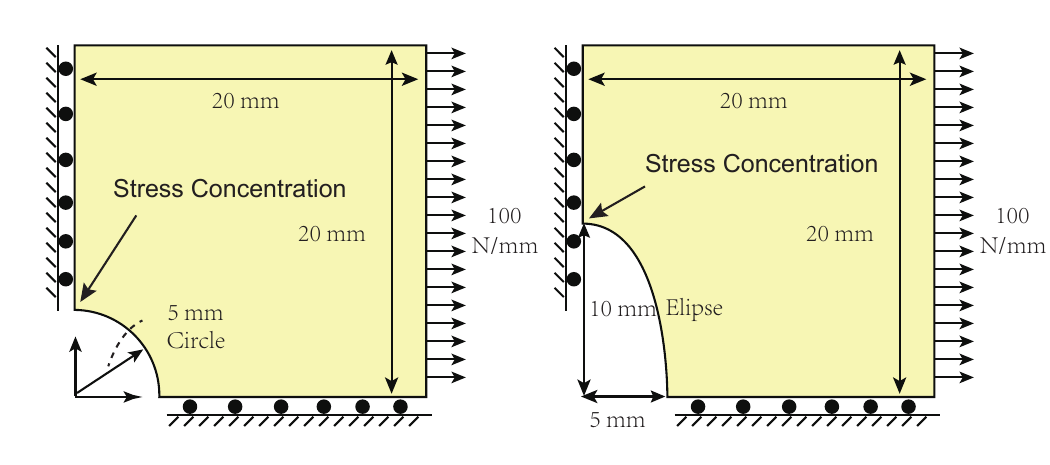}
		\par\end{centering}
	\caption{Schematic of the square plate with a hole: The left figure shows a square plate with a circular hole of radius 5 mm, and the plate has a side length of 20 mm. The load is a uniform load of 100 \(N/mm\) applied to the right boundary of the plate. The right figure shows a square plate with an elliptical hole, where the major and minor axes of the ellipse are 10 mm and 5 mm, respectively. The plate has a side length of 20 mm, and the load is a uniform load of 100 \(N/mm\) applied to the right boundary of the plate. \label{fig:plate_hole}}
\end{figure}

This numerical example explores the generalization capability of transfer learning across different geometries. The example we adopt is a square plate with a hole, as shown in \Cref{fig:plate_hole}. The square plate with a hole is a classic benchmark for stress concentration in solid mechanics, where stress concentration occurs at the edge of the hole. The material is linear elastic, with an elastic modulus of \(E = 1000\) MPa and a Poisson's ratio of \(0.3\). The governing equations are the same as in \Cref{eq:strong_form_elasticity}.

Since this is an elastic mechanics problem, there exists a corresponding energy form of PINNs (DEM), and it can also be solved using the strong form of PINNs \citet{wang2025physics}. Considering that the energy form of PINNs involves too many hyperparameters for solving this elastic mechanics problem, we only use DEM here. DEM requires the essential boundary conditions to be satisfied in advance. The neural network constructs the possible displacement field as:
\begin{equation}
	\begin{aligned}u_{x} & =x\cdot NN(\frac{\boldsymbol{x}}{L};\boldsymbol{\theta})\\
		u_{y} & =y\cdot NN(\frac{\boldsymbol{x}}{L};\boldsymbol{\theta})
	\end{aligned},
	\label{eq:plate_admiss}
\end{equation}
where \(L = 20\) is the side length of the square plate, and \(\boldsymbol{x}/L\) is used for normalization. Since the original size of the input coordinates is much larger than 1 (\(L \gg 1\)), using the original scale would lead to the vanishing gradient problem in the neural network. This occurs because large input values cause the \(\tanh\) activation function to operate in its saturation region, where its derivative becomes very small. As a result, during backpropagation, the gradients of the trainable parameters diminish significantly, causing the learning process to stagnate.
The accuracy of DEM largely depends on the integration scheme. For DEM, we use triangular integration, as it performs better than traditional Monte Carlo integration in DEM \citet{wang2025physics}.

\begin{figure}
	\begin{centering}
		\includegraphics[scale=0.4]{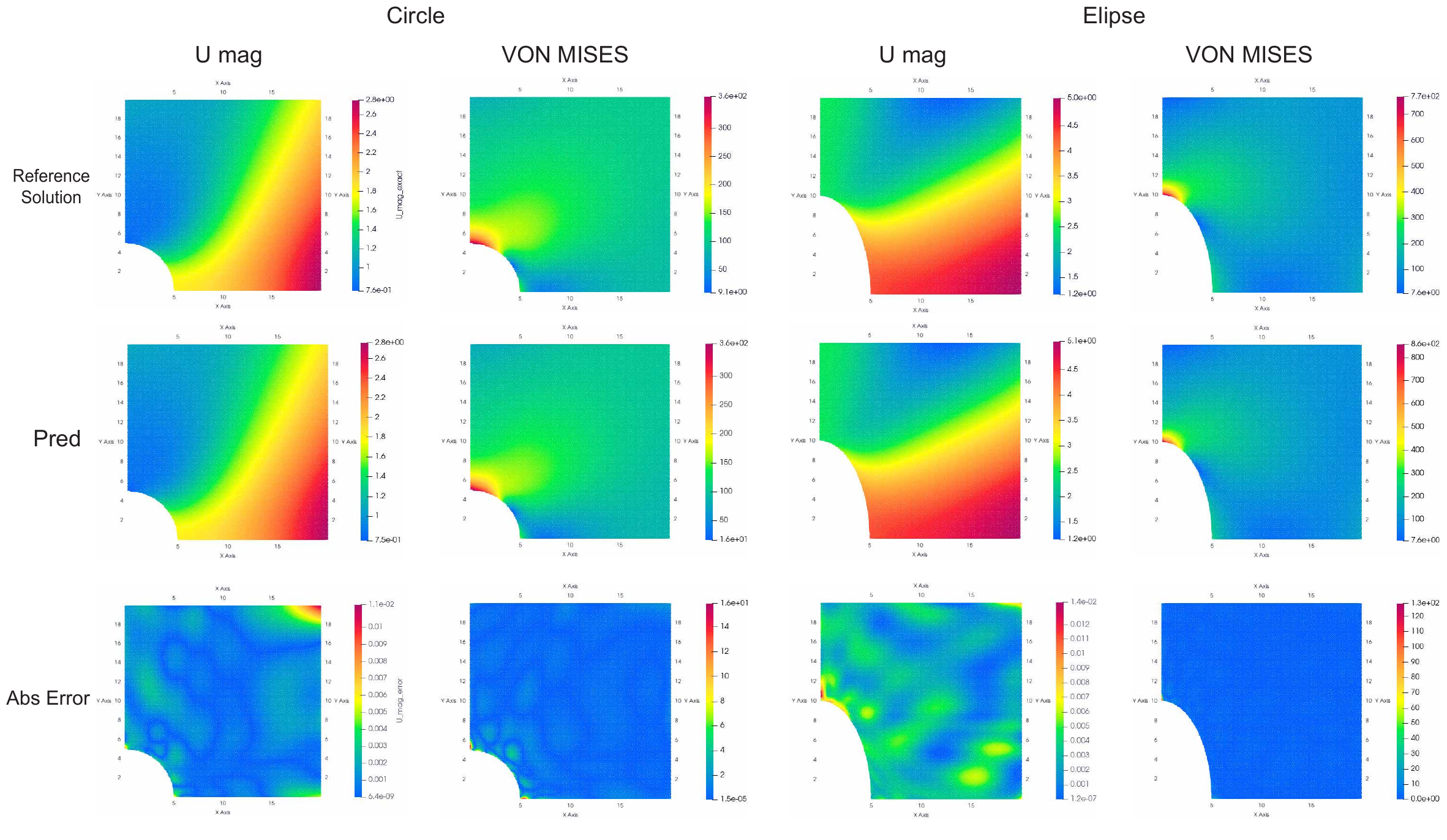}
		\par\end{centering}
	\caption{Predicted displacement field ($u_{mag}$) and Von Mises stress for the square plate with a hole:  The first row shows the reference solution computed using Abaqus, the second row shows the predicted solution from DEM, and the third row shows the absolute error. \label{fig:plate_holes_contourf}}
\end{figure}

\Cref{fig:plate_holes_contourf} shows the predicted displacement field $u_{mag}$ and Von Mises stress results from DEM (after 60,000 epochs), where the displacement field is \(u_{mag} = \sqrt{u_{x}^{2} + u_{y}^{2}}\), and the Von Mises stress is as shown in \Cref{eq:von_mises}. The reference solution is obtained using the finite element software Abaqus. The plate with a circular hole has 18,000 CPS8R elements, and the plate with an elliptical hole has 14,993 CPS8R elements. Both the circular and elliptical cases have undergone finite element convergence analysis, ensuring the reliability of the reference solution. For DEM, the plate with a circular hole and the plate with an elliptical hole use 75,309 and 71,403 uniformly distributed triangular integration points, respectively.

\begin{figure}
	\begin{centering}
		\includegraphics{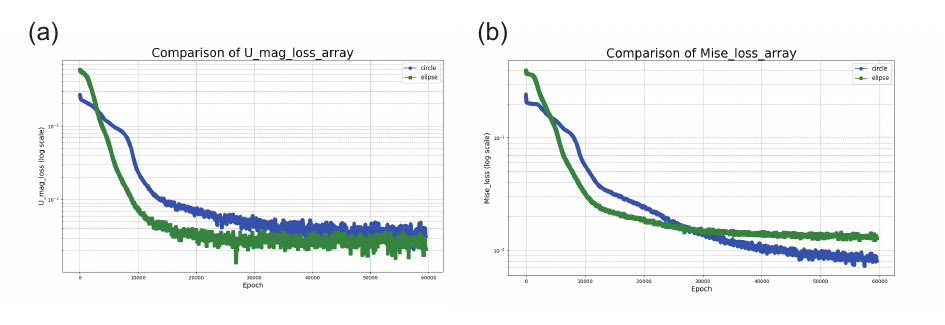}
		\par\end{centering}
	\caption{Evolution of the relative error for the displacement field \(\mathcal{L}_{2}\) $u_{mag}$ (a) and Von Mises stress \(\mathcal{H}_{1}\) (b) for the square plate with a hole. \label{fig:plate_error}}
\end{figure}

To quantify the accuracy of DEM, \Cref{fig:plate_error} shows the evolution of the relative error for DEM in the square plate with a hole problem. The results show that DEM can converge well to the reference solution. Next, we test the effect of transfer learning on geometry generalization.

\begin{figure}
	\begin{centering}
		\includegraphics[scale=0.9]{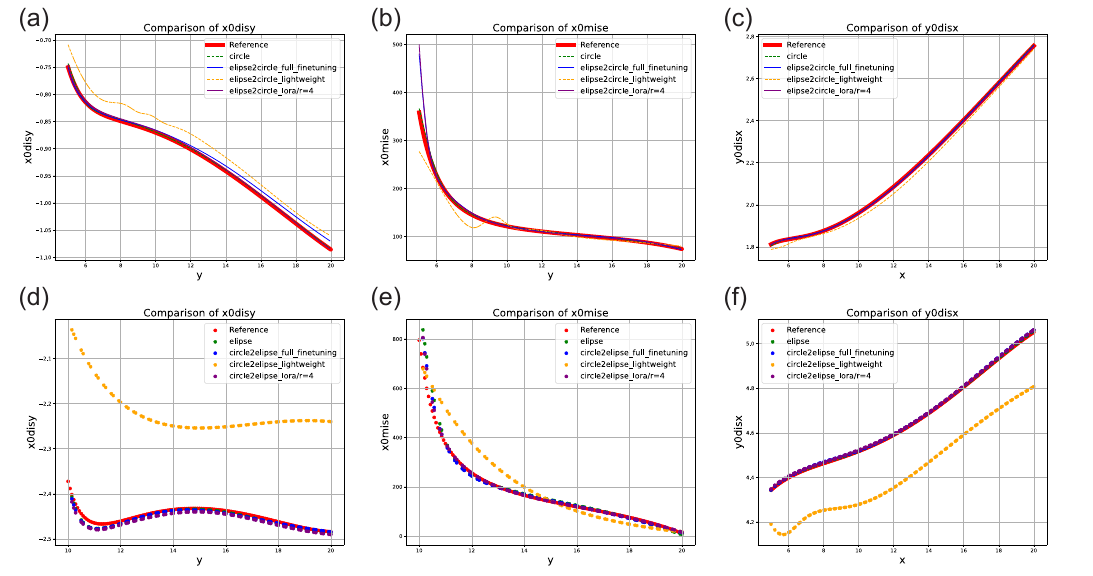}
		\par\end{centering}
	\caption{Performance of transfer learning on the lines \(x = 0\) and \(y = 0\) for the square plate with a hole: (a, b, c) Transfer from the elliptical hole plate to the circular hole plate, where the "circle" curve represents the result without transfer learning. (d, e, f) Transfer from the circular hole plate to the elliptical hole plate, where the "elipse" curve represents the result without transfer learning. The first column shows the \(y\)-direction displacement \(u_{y}\) on the line \(x = 0\); the second column shows the Von Mises stress on the line \(x = 0\); the third column shows the \(x\)-direction displacement \(u_{x}\) on the line \(y = 0\). \label{fig:transfer_line_plate}}
\end{figure}

The characteristic of the square plate with a hole problem is that the maximum \(y\)-direction displacement \(u_{y}\) and Von Mises stress occur on the line \(x = 0\), while the maximum \(x\)-direction displacement \(u_{x}\) occurs on the line \(y = 0\). Therefore, we explore the performance of transfer learning on the lines \(x = 0\) and \(y = 0\). We adopt three transfer learning schemes: full finetuning, lightweight finetuning (only training the last layer), and LoRA with a rank of 4. \Cref{fig:transfer_line_plate} shows the performance of transfer learning on the lines \(x = 0\) and \(y = 0\). The results indicate that lightweight finetuning performs the worst, while full finetuning and LoRA perform the best. Note that the results are obtained after 60,000 epochs for both the source domain and target domain.

\begin{figure}
	\begin{centering}
		\includegraphics{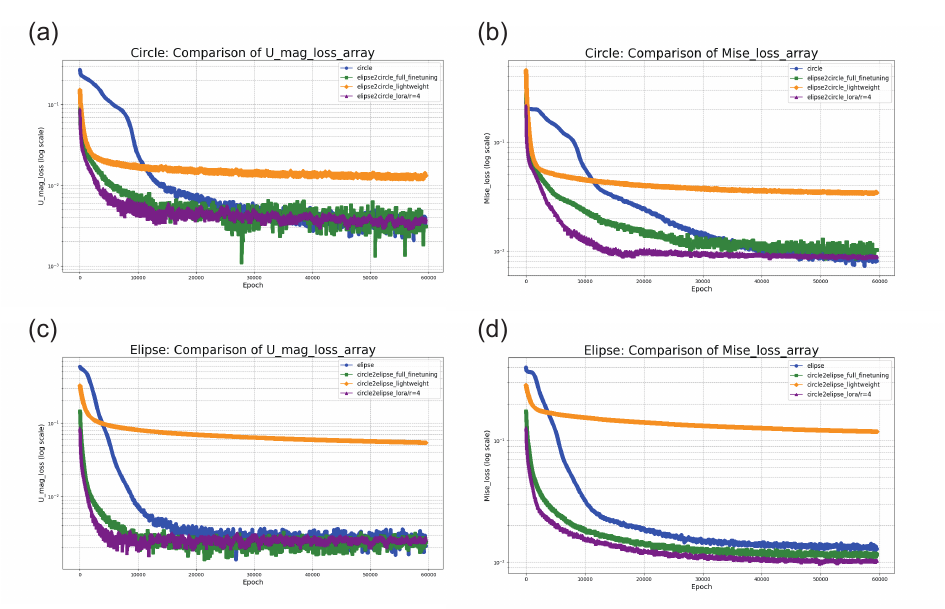}
		\par\end{centering}
	\caption{Evolution of the relative error for displacement ($u_{mag}$) and Von Mises stress under different transfer learning schemes for the square plate with a hole: (a) Evolution of the relative error for the displacement field of the circular hole plate. (b) Evolution of the relative error for the Von Mises stress of the circular hole plate. (c) Evolution of the relative error for the displacement field of the elliptical hole plate. (d) Evolution of the relative error for the Von Mises stress of the elliptical hole plate. (a, b) The source domain is the elliptical hole plate, and the target domain is the circular hole plate. (c, d) The source domain is the circular hole plate, and the target domain is the elliptical hole plate. \label{fig:transfer_learning_l2_plate}}
\end{figure}

To quantify the accuracy of transfer learning, \Cref{fig:transfer_learning_l2_plate} shows the evolution of the relative error for different transfer learning schemes. The results demonstrate that transfer learning can significantly improve the convergence speed. \Cref{tab:transfer_circle_elipse} provides the specific efficiency and accuracy of different transfer learning schemes. The results show that LoRA achieves the highest accuracy, although the improvement is not significant.

\begin{table}
	\caption{Accuracy and efficiency of different transfer learning schemes for geometry generalization: The relative errors \(\mathcal{L}_{2}\) ($u_{mag}$) and \(\mathcal{H}_{1}\) (Von Mises) are measured after 60,000 iterations. \(Circle \rightarrow Elipse\) refers to the case where the source domain is the plate with a circular hole, and the target domain is the plate with an elliptical hole. The reverse is also true. \label{tab:transfer_circle_elipse}}
	\begin{adjustbox}{max width=\textwidth}
	\begin{centering}
	\begin{tabular}{ccccccc}
		\toprule 
		& No transfer & Full\_finetuning & Lightweight\_finetuning (last) & LoRA (\(r=1\)) & LoRA (\(r=4\)) & LoRA (\(r=100\))\tabularnewline
		\midrule
		Time (s, 1000 Epochs) & 46.82 & 46.82 & \textbf{43.48} & 46.53 & 46.77 & 46.80\tabularnewline
		Trainable parameters & 30802 & 30802 & \textbf{202} & 900 & 2700 & 60300\tabularnewline
		Relative error \(\mathcal{L}_{2}\) for displacement: \(Circle \rightarrow Elipse\) & 0.0021302 & 0.0023966 & 0.054191 & \textbf{0.0020030} & 0.0026557 & 0.0024090\tabularnewline
		Relative error \(\mathcal{L}_{2}\) for displacement: \(Elipse \rightarrow Circle\) & 0.0031061 & 0.0031083 & 0.013176 & 0.0031666 & 0.0038253 & \textbf{0.00086261}\tabularnewline
		Relative error \(\mathcal{H}_{1}\) for stress: \(Circle \rightarrow Elipse\) & 0.012843 & 0.011563 & 0.11855 & 0.012162 & \textbf{0.010285} & 0.012158\tabularnewline
		Relative error \(\mathcal{H}_{1}\) for stress: \(Elipse \rightarrow Circle\) & 0.0080823 & 0.010230 & 0.034274 & 0.0093254 & 0.0088848 & \textbf{0.0080786}\tabularnewline
		\bottomrule
	\end{tabular}
\par\end{centering}
\end{adjustbox}
\end{table}

\section{Discussion \label{sec:Dicussion}}

\subsection{The Rank in LoRA \label{subsec:The-rank-in-LoRA}}

\begin{figure}
	\begin{centering}
		\includegraphics[scale=0.7]{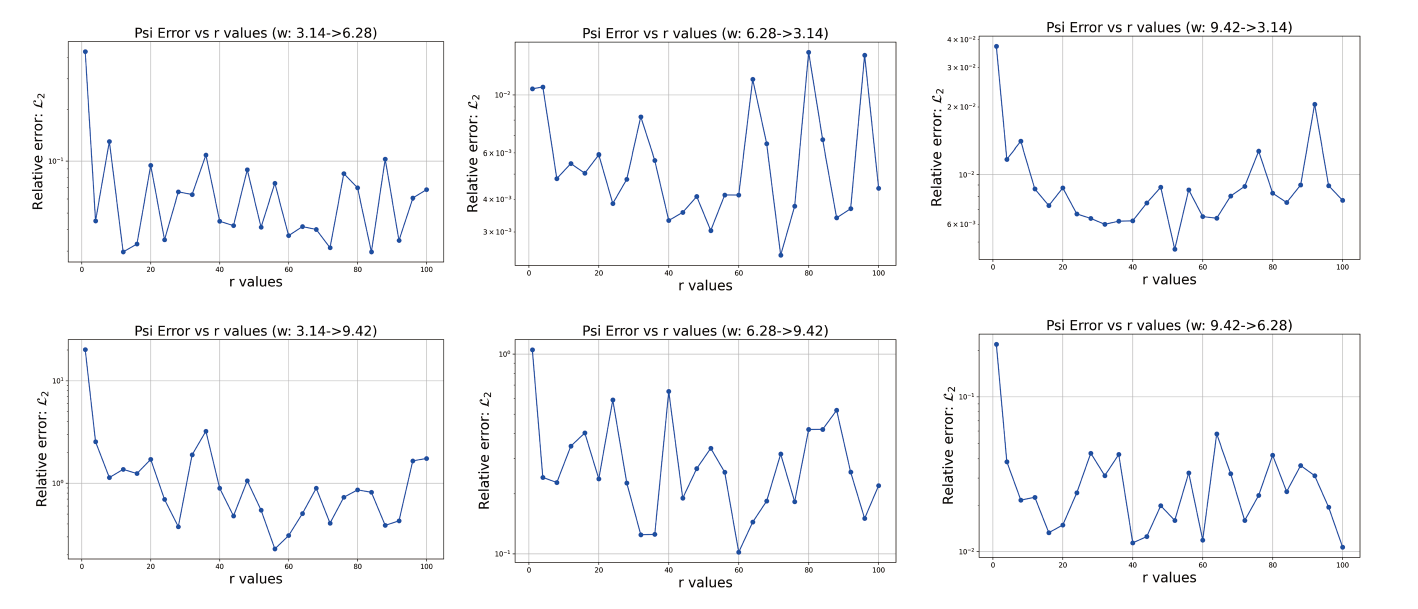}
		\par\end{centering}
	\caption{Performance of different ranks in LoRA for PINNs: We vary the rank in LoRA as \(1, 4, 8, 12, \dots, 96, 100\) and observe the relative error \(\mathcal{L}_{2}\) in \ref{subsec:Transfer_boundary}. \(w = X \rightarrow Y\) indicates that PINNs are pre-trained on \(w = X\) as the source domain and then fine-tuned on \(w = Y\) as the target domain. The relative error \(\mathcal{L}_{2}\) is averaged over the last 1,000 iterations. \label{fig:LoRA_rank}}
\end{figure}

Theoretically, LoRA is a well-suited technique for fine-tuning PINNs, and it serves as a more generalized form of full finetuning and lightweight finetuning. However, in this manuscript, all experiments assume a predetermined rank for LoRA, with most cases using \(r = 4\). Therefore, finding an appropriate method to determine the rank in LoRA is crucial. \Cref{fig:LoRA_rank} illustrates the results of different ranks in LoRA for the case described in \Cref{subsec:Transfer_boundary}. LoRA is applied only to the fully connected structure \([3,100,100,100,100,2]\), training \([100,100,100,100]\), with all LoRA layers sharing the same rank. It is evident that the larger the discrepancy between the source domain and the target domain, the higher the optimal rank. For example, when the source domain is \(\pi\) and the target domain is \(2\pi\), the optimal rank is 12. However, when the source domain is \(\pi\) and the target domain is \(3\pi\), the optimal rank increases to 56. This suggests that the rank in LoRA can be determined based on the similarity between the source and target domains.

\begin{figure}
	\begin{centering}
		\includegraphics[scale=0.55]{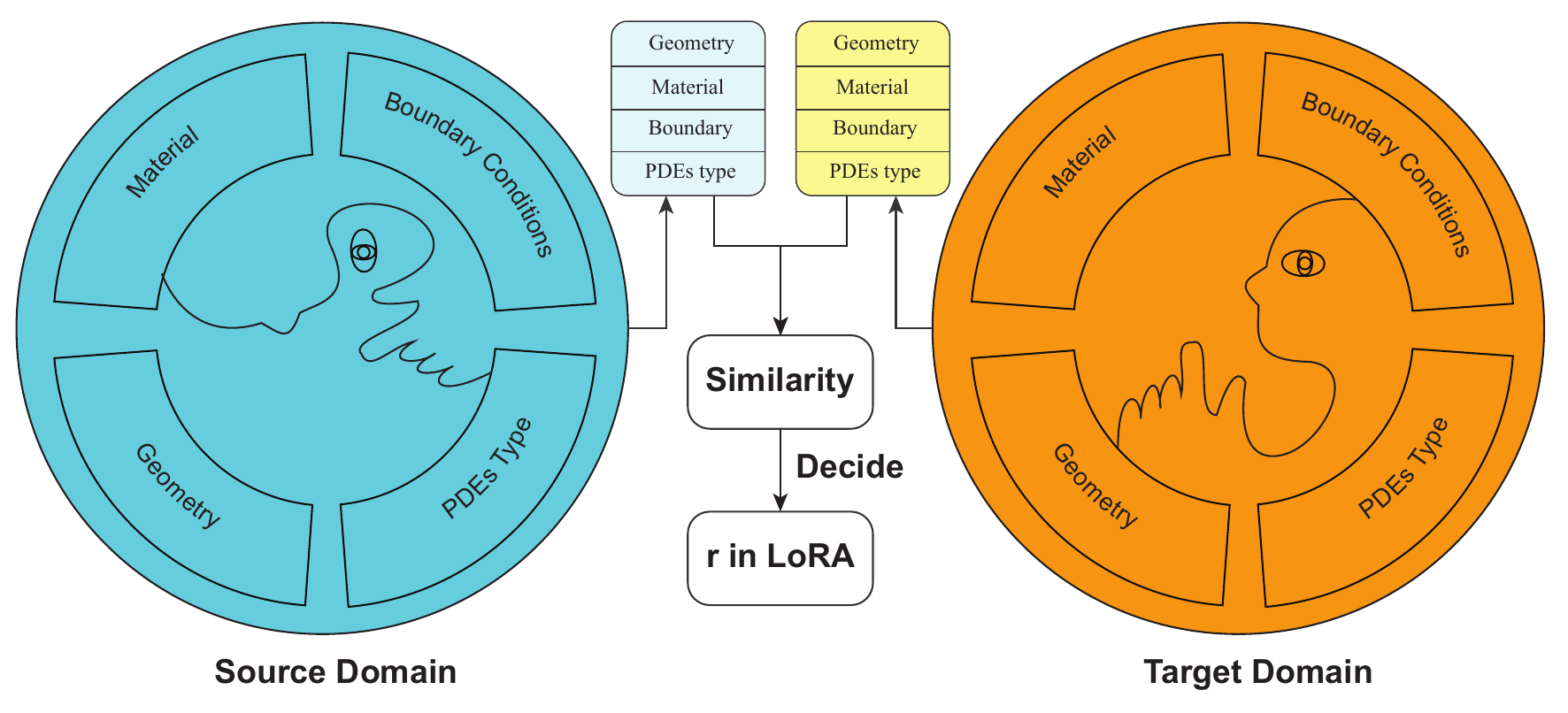}
		\par\end{centering}
	\caption{Schematic for determining the rank \(r\) in LoRA \label{fig:decide_r_in_LoRA}}
\end{figure}

The similarity can be determined based on geometry, material, boundary conditions, and PDEs types, as shown in \Cref{fig:decide_r_in_LoRA}. For instance, we can convert geometry, material, boundary conditions, and PDEs types into feature vectors \(\boldsymbol{V}\) and compute the similarity using cosine similarity:
\begin{equation}
	similarity = \frac{\boldsymbol{V}_{s} \cdot \boldsymbol{V}_{t}}{||\boldsymbol{V}_{s}|| \cdot ||\boldsymbol{V}_{t}||}\label{eq:similiar_cos},
\end{equation}
where \(\boldsymbol{V}_{s}\) is the feature vector of the source domain, and \(\boldsymbol{V}_{t}\) is the feature vector of the target domain. Note that the feature vectors are determined by geometry, material, boundary conditions, and PDEs types.

This work does not explore how to automatically determine the rank in LoRA, particularly how to characterize geometry, material, boundary conditions, and PDEs types into an abstract vector \(\boldsymbol{V}\) shown in \Cref{eq:similiar_cos}. Additionally, all layers share the same rank, without adaptive variations for each layer. In the future, this remains a promising direction for transfer learning in AI for PDEs.

\subsection{Feature Fusion in Transfer Learning}

\begin{figure}
	\begin{centering}
		\includegraphics[scale=0.4]{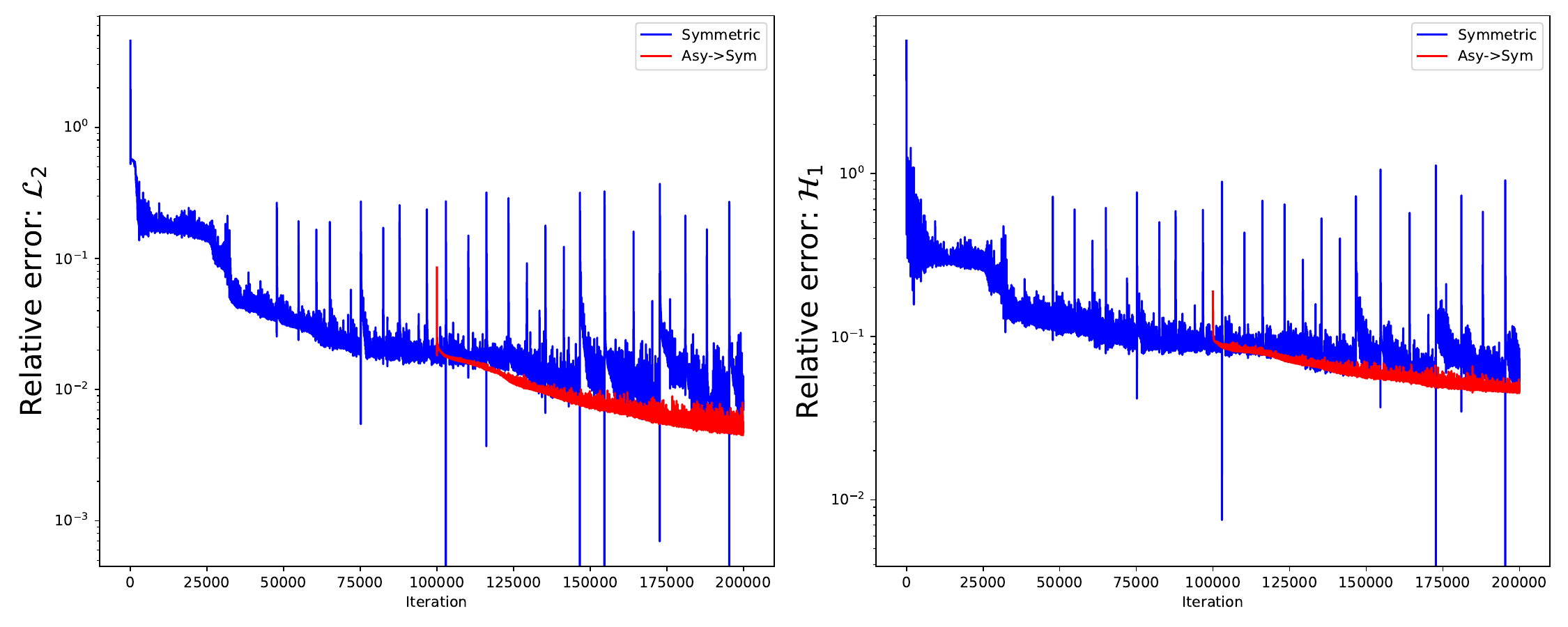}
		\par\end{centering}
	\caption{Scenarios showing the evolution of relative errors for displacement (left) and stress (right): Symmetric refers to training without transfer learning for 200,000 iterations. Asy $\rightarrow$ Sym refers to pre-training on the asymmetric porosity distribution (source domain) for 100,000 iterations, followed by training on the symmetric porosity distribution (target domain) for another 100,000 iterations.
Note that the total number of iterations is the same: 100,000 in the source domain and 100,000 in the target domain and non-transfer learning 200,000 iterations.\label{fig:Scenarios-fused_result_beam}}
\end{figure}

We observed an interesting phenomenon in our experiments: learning across different scenarios can improve the accuracy of PINNs. For example, as shown in \Cref{subsec:Transfer-learning-material}, \Cref{fig:Scenarios-fused_result_beam} demonstrates that training for 100,000 iterations in the source domain followed by 100,000 iterations in the target domain yields higher accuracy than training for 200,000 iterations without transfer learning. Note that the total number of iterations remains the same for both cases (100,000 in the source domain and 100,000 in the target domain for transfer learning, versus 200,000 iterations without transfer learning).

\begin{figure}
	\begin{centering}
		\includegraphics[scale=0.6]{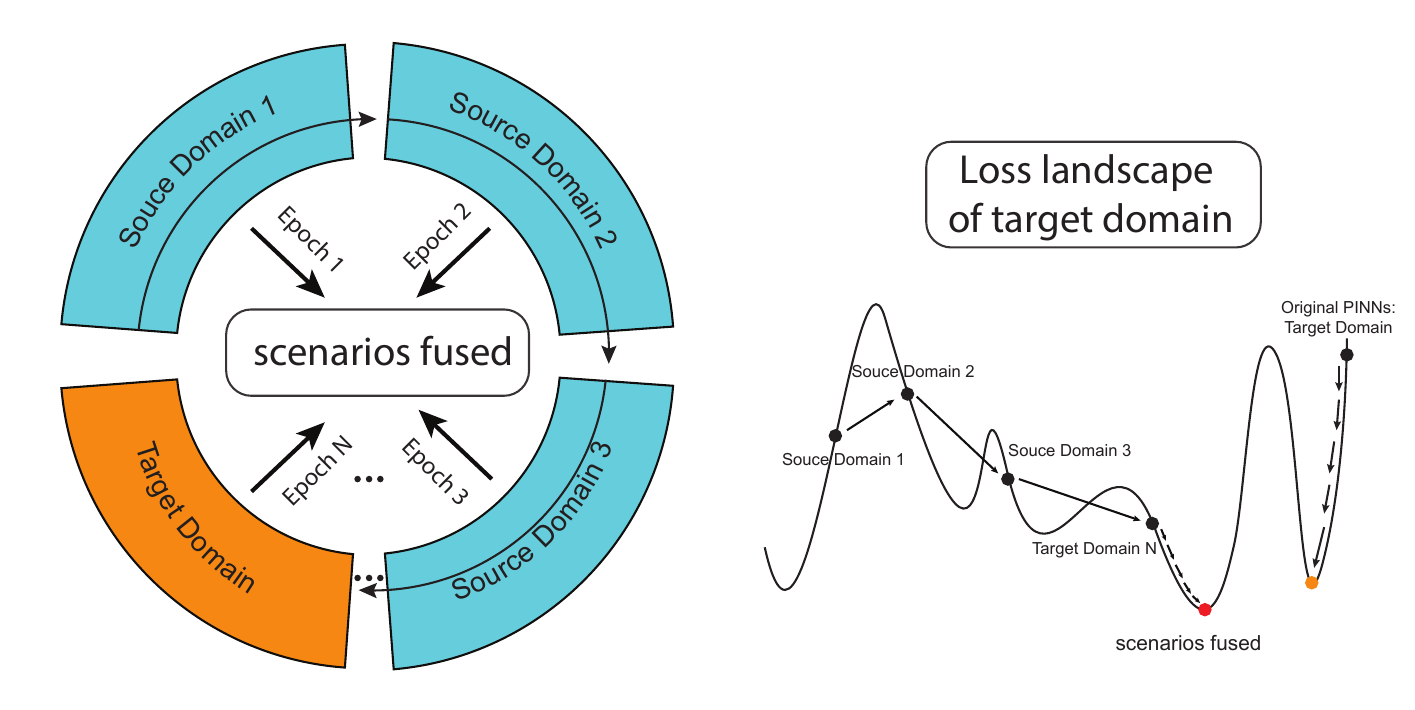}
		\par\end{centering}
	\caption{Schematic of scenario fusion: The left figure divides the total iterations required for PINNs into \(N\) scenarios, with the required iterations being \(E_{1}\), \(E_{2}\), \(\dots\), \(E_{N-1}\), and \(E_{N}\). The last scenario \(N\) is the target domain, while the first \(N-1\) scenarios are sequentially trained as source domains. Note that we control \(E_{PINNs} = \sum_{i=1}^{N} E_{i}\). The right figure shows the loss function landscape of the target domain. By leveraging different source domains, the optimization process resembles jumping, allowing it to bypass many local optima. \label{fig:Scenarios-fused}}
\end{figure}

This indicates that fusing different scenarios can enhance the performance of PINNs. In the future, exploring scenario fusion in PINNs, as illustrated in \Cref{fig:Scenarios-fused}, could further improve PINNs accuracy. We emphasize that the total number of iterations remains unchanged, so efficiency is preserved while accuracy improves. Future work should investigate which scenario fusions enhance PINNs' accuracy and how many scenarios should be fused.

\section{Conclusion \label{sec:Conclusion}}

We investigated the application of transfer learning in PINNs, focusing on two key forms: the strong form and the energy form. Considering that PINNs require retraining when faced with new boundary conditions, materials, or geometries, we explored the effectiveness of transfer learning across various cases, including generalization to different boundary conditions, materials, and geometries. The results demonstrate that LoRA and full finetuning significantly improve convergence speed and accuracy in most cases, while lightweight finetuning performs poorly in most scenarios. Additionally, the rank in LoRA can be determined based on the similarity between the source and target domains.

Although we validated the performance of transfer learning in PINNs, AI for PDEs \citet{yizheng2024ai,wang2024artificial} includes two other key approaches: operator learning \citet{li2020fourier,DeepOnet} and Physics-Informed Neural Operators (PINO) \citet{li2024physics,eshaghi2024variational}. In the future, we will further explore the effectiveness of transfer learning in operator learning and PINO within AI for PDEs. Moreover, determining the low-rank approximation in LoRA is crucial. While we discovered that the optimal rank range in LoRA is related to the similarity between domains, future work should develop methods to effectively determine the rank in LoRA and adaptively adjust the rank for different hidden layers based on the similarity between the source and target domains. Leveraging Green's functions, which allow rapid integration of boundary conditions to obtain solutions, transfer learning can be applied to generalize and accelerate PINNs computations across different scenarios.

We believe that transfer learning is a key technology for future large-scale computational mechanics models. For example, the recently proposed ICON \citet{yang2023context} integrates multiple operators. In the future, transfer learning will play a particularly important role in fine-tuning pre-trained large-scale computational mechanics models on smaller, personalized datasets.

\section*{Declaration of competing interest}
The authors declare that they have no known competing financial interests or personal relationships that could
have appeared to influence the work reported in this paper.

\section*{Acknowledgement}
The study was supported by the Key Project of the National Natural Science Foundation of China (12332005) and scholarship from Bauhaus University in Weimar.

\appendix

\section{Taylor Green Vortex boundary and initial conditions setup\label{sec:Taylor-Green-Vortex_boundary_initial}}

For the Taylor Green Vortex, we apply the initial and boundary conditions based on the analytical solution \Cref{eq:exact_solution_taylor}. The initial conditions are as follows:
\begin{equation}
	\begin{aligned}
		\psi & = \frac{1}{w}\cos(wx)\cos(wy) \\
		\Omega & = -2w\cos(wx)\cos(wy) \\
		u & = -\cos(wx)\sin(wy) \\
		v & = \sin(wx)\cos(wy) \\
		p & = -\frac{1}{2}[\cos^{2}(wx)+\cos^{2}(wy)]
	\end{aligned}.
	\label{eq:initial_taylor}
\end{equation}

Our simulation domain is $[0,1]^{2}$, so we obtain the boundary conditions as:
\begin{equation}
	\begin{cases}
		\psi = \frac{1}{w}\exp\left(-\frac{2w^{2}}{Re}t\right)\cos(wx) & , x \in [0,1], y = 0 \\
		\Omega = -2w\exp\left(-\frac{2w^{2}}{Re}t\right)\cos(wx) & , x \in [0,1], y = 0 \\
		u = 0 & , x \in [0,1], y = 0 \\
		v = \exp\left(-\frac{2w^{2}}{Re}t\right)\sin(wx) & , x \in [0,1], y = 0
	\end{cases},
\end{equation}

\begin{equation}
	\begin{cases}
		\psi = \frac{1}{w}\exp\left(-\frac{2w^{2}}{Re}t\right)\cos(wx)\cos(w) & , x \in [0,1], y = 1 \\
		\Omega = -2w\exp\left(-\frac{2w^{2}}{Re}t\right)\cos(wx)\cos(w) & , x \in [0,1], y = 1 \\
		u = -\exp\left(-\frac{2w^{2}}{Re}t\right)\cos(wx)\sin(w) & , x \in [0,1], y = 1 \\
		v = \exp\left(-\frac{2w^{2}}{Re}t\right)\sin(wx)\cos(w) & , x \in [0,1], y = 1
	\end{cases},
\end{equation}

\begin{equation}
	\begin{cases}
		\psi = \frac{1}{w}\exp\left(-\frac{2w^{2}}{Re}t\right)\cos(wy) & , x = 0, y \in [0,1] \\
		\Omega = -2w\exp\left(-\frac{2w^{2}}{Re}t\right)\cos(wy) & , x = 0, y \in [0,1] \\
		u = -\exp\left(-\frac{2w^{2}}{Re}t\right)\sin(wy) & , x = 0, y \in [0,1] \\
		v = 0 & , x = 0, y \in [0,1]
	\end{cases},
\end{equation}

\begin{equation}
	\begin{cases}
		\psi = \frac{1}{w}\exp\left(-\frac{2w^{2}}{Re}t\right)\cos(w)\cos(wy) & , x = 1, y \in [0,1] \\
		\Omega = -2w\exp\left(-\frac{2w^{2}}{Re}t\right)\cos(w)\cos(wy) & , x = 1, y \in [0,1] \\
		u = -\exp\left(-\frac{2w^{2}}{Re}t\right)\cos(w)\sin(wy) & , x = 1, y \in [0,1] \\
		v = \exp\left(-\frac{2w^{2}}{Re}t\right)\sin(w)\cos(wy) & , x = 1, y \in [0,1]
	\end{cases}.
\end{equation}

\section{Computation graph of PINNs\label{sec:Computation-graph-of_PINN}}

We found that the efficiency of lightweight finetuning did not significantly improve compared to full finetuning. This is because PINNs require differentiation with respect to the input, which approximates the differential operator. This step creates new computation graphs. To illustrate the issue, let us consider a simple PDE:
\begin{equation}
	\nabla^{2}T=0 \label{eq:computational_graph}.
\end{equation}

\begin{figure}
	\begin{centering}
		\includegraphics[scale= 0.42]{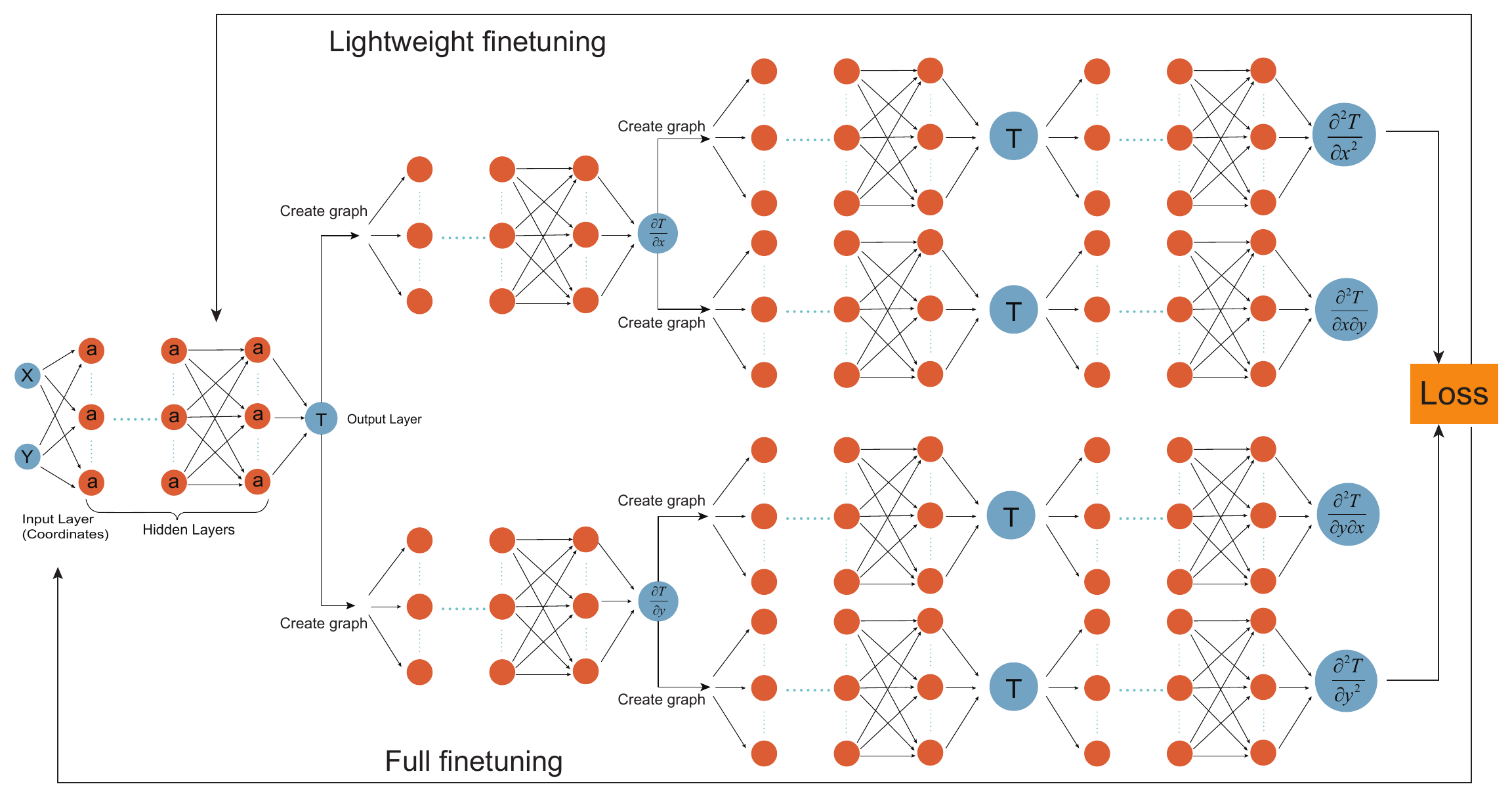}
		\par\end{centering}
	\caption{Computation graph of PINNs illustration\label{fig:Computation-graph}}
\end{figure}

\Cref{fig:Computation-graph} shows an explanation of the computation graph for PINNs. When constructing the differential operator, new computation graphs are generated. Consider the PDEs in \Cref{eq:computational_graph}, where the neural network is $T = NN(\boldsymbol{x}; \boldsymbol{\theta})$, with $T$ as the network output, $\boldsymbol{x}$ as the input coordinates, and $\boldsymbol{\theta}$ as the trainable parameters. If the computation graph for $\boldsymbol{x} \to T$ has $C_{T}$ steps, then the computation graph for $\boldsymbol{x} \to \partial T/\partial \boldsymbol{x}$ has $2C_{T}$ steps, and the computation graph for $\boldsymbol{x} \to \partial^{2}T/\partial \boldsymbol{x}^{2}$ has $4C_{T}$ steps. Therefore, if the highest derivative order of the PDEs is $M$, the computation graph will have a maximum of $2^{M}C$ steps, where $C$ is the basic number of steps for the network's computation from coordinates to the target field.

Thus, for full finetuning, the maximum computation graph has $2^{M}C$ steps; for lightweight finetuning, where only the last layer is trained, the maximum computation graph has $(2^{M}-1)C + 1$ steps. The ratio of computation graphs between lightweight finetuning, $C_{light}$, and full finetuning, $C_{full}$, is given by:

\begin{equation}
	\frac{C_{light}}{C_{full}} = \frac{(2^{M}-1)C+1}{2^{M}C} = 1 - \frac{C-1}{2^{M}C} \label{eq:computational_ratio}.
\end{equation}

As a result, the efficiency gain of lightweight finetuning during backpropagation, when computing gradients for the optimizable parameters, decreases as the highest derivative order $M$ of the PDEs increases.

In conclusion, PINNs inherently involve the computation of differential operators, so when using PyTorch's Automatic Differentiation \citep{automatic_differential}, more computation graphs are inevitably created. Therefore, during backpropagation, the number of graphs that lightweight finetuning must traverse is not significantly smaller than that of full finetuning.

\section{Computation of LoRA\label{sec:Computation-of-LoRA}}

Let's recall the parameters of LoRA:
\begin{equation}
	\boldsymbol{W}^{*} = \boldsymbol{W} + \alpha \boldsymbol{A} \boldsymbol{B}.
\end{equation}

When LoRA is not used, the gradient of the trainable parameters is:
\begin{equation}
	\frac{\partial \mathcal{L}}{\partial \boldsymbol{W}} = \frac{\partial \mathcal{L}}{\partial W_{ij}}.
\end{equation}

When considering LoRA, the gradient of the trainable parameters is:
\begin{equation}
	\begin{aligned}
		\frac{\partial \mathcal{L}}{\partial \boldsymbol{A}} & = \frac{\partial \mathcal{L}}{\partial \boldsymbol{W}^{*}} \frac{\partial \boldsymbol{W}^{*}}{\partial \boldsymbol{A}} = \frac{\partial \mathcal{L}}{\partial W_{ij}^{*}} \frac{\partial (W_{ij} + \alpha A_{im} B_{mj})}{\partial A_{kl}} \\
		& = \frac{\partial \mathcal{L}}{\partial W_{ij}^{*}} \alpha \delta_{ik} \delta_{ml} B_{mj} = \alpha \frac{\partial \mathcal{L}}{\partial W_{kj}^{*}} B_{lj} \\
		\frac{\partial \mathcal{L}}{\partial \boldsymbol{B}} & = \frac{\partial \mathcal{L}}{\partial \boldsymbol{W}^{*}} \frac{\partial \boldsymbol{W}^{*}}{\partial \boldsymbol{B}} = \frac{\partial \mathcal{L}}{\partial W_{ij}^{*}} \frac{\partial (W_{ij} + \alpha A_{im} B_{mj})}{\partial B_{kl}} \\
		& = \frac{\partial \mathcal{L}}{\partial W_{ij}^{*}} \alpha A_{im} \delta_{mk} \delta_{jl} = \alpha \frac{\partial \mathcal{L}}{\partial W_{il}^{*}} A_{ik}
	\end{aligned}.
	\label{eq:LoRA_computation}
\end{equation}
Since $\boldsymbol{W}^{*}$ and $\boldsymbol{W}$ have the same shape, the computation cost of $\frac{\partial \mathcal{L}}{\partial W_{ij}}$ is the same as $\frac{\partial \mathcal{L}}{\partial W_{ij}^{*}}$.  \Cref{eq:LoRA_computation} shows that LoRA introduces an additional matrix multiplication, which means that while the number of trainable parameters is reduced in LoRA, the internal computation cost for calculating the gradient of trainable parameters is actually increased compared to not using LoRA.

This leads to the longer computation time of LoRA compared to full finetuning, as shown in \Cref{tab:PINNs_strong_transfer}.

\section{Supplementary code}
The code of this work will be available at \url{https://github.com/yizheng-wang/Research-on-Solving-Partial-Differential-Equations-of-Solid-Mechanics-Based-on-PINN} after accepted.

\bibliographystyle{elsarticle-num}
\addcontentsline{toc}{section}{\refname}\bibliography{reference.bib}

\begin{thebibliography}{10}
\expandafter\ifx\csname url\endcsname\relax
  \def\url#1{\texttt{#1}}\fi
\expandafter\ifx\csname urlprefix\endcsname\relax\def\urlprefix{URL }\fi
\expandafter\ifx\csname href\endcsname\relax
  \def\href#1#2{#2} \def\path#1{#1}\fi

\bibitem{loss_is_minimum_potential_energy}
E.~Samaniego, C.~Anitescu, S.~Goswami, V.~M. Nguyen-Thanh, H.~Guo, K.~Hamdia,
  X.~Zhuang, T.~Rabczuk, An energy approach to the solution of partial
  differential equations in computational mechanics via machine learning:
  Concepts, implementation and applications, Computer Methods in Applied
  Mechanics and Engineering 362 (2020) 112790.

\bibitem{PINN_review}
G.~E. Karniadakis, I.~G. Kevrekidis, L.~Lu, P.~Perdikaris, S.~Wang, L.~Yang,
  Physics-informed machine learning, Nature Reviews Physics 3~(6) (2021)
  422--440.
\newblock \href {https://doi.org/10.1038/s42254-021-00314-5}
  {\path{doi:10.1038/s42254-021-00314-5}}.

\bibitem{wang2021learning}
S.~Wang, H.~Wang, P.~Perdikaris, Learning the solution operator of parametric
  partial differential equations with physics-informed deeponets, Science
  advances 7~(40) (2021) eabi8605.

\bibitem{wang2024artificial}
Y.~Wang, J.~Bai, Z.~Lin, Q.~Wang, C.~Anitescu, J.~Sun, M.~S. Eshaghi, Y.~Gu,
  X.-Q. Feng, X.~Zhuang, et~al., Artificial intelligence for partial
  differential equations in computational mechanics: A review, arXiv preprint
  arXiv:2410.19843 (2024).

\bibitem{yizheng2024ai}
W.~Yizheng, Z.~Xiaoying, T.~Rabczuk, L.~Yinghua, Ai for pdes in solid
  mechanics: A review, Advances in Mechanics 54~(3) (2024) 1--57.

\bibitem{PINN_original_paper}
M.~Raissi, P.~Perdikaris, G.~E. Karniadakis, Physics-informed neural networks:
  A deep learning framework for solving forward and inverse problems involving
  nonlinear partial differential equations, Journal of Computational Physics
  378 (2019) 686--707.

\bibitem{DeepOnet}
L.~Lu, P.~Jin, G.~Pang, Z.~Zhang, G.~E. Karniadakis, Learning nonlinear
  operators via deeponet based on the universal approximation theorem of
  operators, Nature Machine Intelligence 3~(3) (2021) 218--229.
\newblock \href {https://doi.org/10.1038/s42256-021-00302-5}
  {\path{doi:10.1038/s42256-021-00302-5}}.

\bibitem{li2020fourier}
Z.~Li, N.~Kovachki, K.~Azizzadenesheli, B.~Liu, K.~Bhattacharya, A.~Stuart,
  A.~Anandkumar, Fourier neural operator for parametric partial differential
  equations, arXiv preprint arXiv:2010.08895 (2020).

\bibitem{li2024physics}
Z.~Li, H.~Zheng, N.~Kovachki, D.~Jin, H.~Chen, B.~Liu, K.~Azizzadenesheli,
  A.~Anandkumar, Physics-informed neural operator for learning partial
  differential equations, ACM/JMS Journal of Data Science 1~(3) (2024) 1--27.

\bibitem{eshaghi2024variational}
M.~S. Eshaghi, C.~Anitescu, M.~Thombre, Y.~Wang, X.~Zhuang, T.~Rabczuk,
  Variational physics-informed neural operator (vino) for solving partial
  differential equations, arXiv preprint arXiv:2411.06587 (2024).

\bibitem{yang2023context}
L.~Yang, S.~Liu, T.~Meng, S.~J. Osher, In-context operator learning with data
  prompts for differential equation problems, Proceedings of the National
  Academy of Sciences 120~(39) (2023) e2310142120.

\bibitem{desai2021one}
S.~Desai, M.~Mattheakis, H.~Joy, P.~Protopapas, S.~Roberts, One-shot transfer
  learning of physics-informed neural networks, arXiv preprint arXiv:2110.11286
  (2021).

\bibitem{gao2022svd}
Y.~Gao, K.~C. Cheung, M.~K. Ng, Svd-pinns: Transfer learning of
  physics-informed neural networks via singular value decomposition, in: 2022
  IEEE Symposium Series on Computational Intelligence (SSCI), IEEE, 2022, pp.
  1443--1450.

\bibitem{zhuang2020comprehensive}
F.~Zhuang, Z.~Qi, K.~Duan, D.~Xi, Y.~Zhu, H.~Zhu, H.~Xiong, Q.~He, A
  comprehensive survey on transfer learning, Proceedings of the IEEE 109~(1)
  (2020) 43--76.

\bibitem{xu2023transfer}
C.~Xu, B.~T. Cao, Y.~Yuan, G.~Meschke, Transfer learning based physics-informed
  neural networks for solving inverse problems in engineering structures under
  different loading scenarios, Computer Methods in Applied Mechanics and
  Engineering 405 (2023) 115852.

\bibitem{guo2022analysis}
H.~Guo, X.~Zhuang, P.~Chen, N.~Alajlan, T.~Rabczuk, Analysis of
  three-dimensional potential problems in non-homogeneous media with
  physics-informed deep collocation method using material transfer learning and
  sensitivity analysis, Engineering with Computers 38~(6) (2022) 5423--5444.

\bibitem{chakraborty2022domain}
A.~Chakraborty, C.~Anitescu, X.~Zhuang, T.~Rabczuk, Domain adaptation based
  transfer learning approach for solving pdes on complex geometries,
  Engineering with Computers 38~(5) (2022) 4569--4588.

\bibitem{chen2021transfer}
X.~Chen, C.~Gong, Q.~Wan, L.~Deng, Y.~Wan, Y.~Liu, B.~Chen, J.~Liu, Transfer
  learning for deep neural network-based partial differential equations
  solving, Advances in Aerodynamics 3 (2021) 1--14.

\bibitem{PINN_solid_mechanics}
E.~Haghighat, M.~Raissi, A.~Moure, H.~Gomez, R.~Juanes, A physics-informed deep
  learning framework for inversion and surrogate modeling in solid mechanics,
  Computer Methods in Applied Mechanics and Engineering 379 (2021) 113741.
\newblock \href {https://doi.org/10.1016/j.cma.2021.113741}
  {\path{doi:10.1016/j.cma.2021.113741}}.

\bibitem{goswami2020transfer}
S.~Goswami, C.~Anitescu, S.~Chakraborty, T.~Rabczuk, Transfer learning enhanced
  physics informed neural network for phase-field modeling of fracture,
  Theoretical and Applied Fracture Mechanics 106 (2020) 102447.

\bibitem{chakraborty2021transfer}
S.~Chakraborty, Transfer learning based multi-fidelity physics informed deep
  neural network, Journal of Computational Physics 426 (2021) 109942.

\bibitem{hu2021lora}
E.~J. Hu, Y.~Shen, P.~Wallis, Z.~Allen-Zhu, Y.~Li, S.~Wang, L.~Wang, W.~Chen,
  Lora: Low-rank adaptation of large language models, arXiv preprint
  arXiv:2106.09685 (2021).

\bibitem{majumdar2023hyperlora}
R.~Majumdar, V.~Jadhav, A.~Deodhar, S.~Karande, L.~Vig, V.~Runkana, Hyperlora
  for pdes, arXiv preprint arXiv:2308.09290 (2023).

\bibitem{cho2023hypernetwork}
W.~Cho, K.~Lee, D.~Rim, N.~Park, Hypernetwork-based meta-learning for low-rank
  physics-informed neural networks, Advances in Neural Information Processing
  Systems 36 (2023) 11219--11231.

\bibitem{the_foundation_of_solid_mechanics_feng}
Y.~Fung, Foundations of solid mechanics. 1965, Englewood Cliffs, NJ 436 (2010).

\bibitem{wang2025physics}
Y.~Wang, J.~Sun, J.~Bai, C.~Anitescu, M.~S. Eshaghi, X.~Zhuang, T.~Rabczuk,
  Y.~Liu, Kolmogorov arnold informed neural network: A physics-informed deep
  learning framework for solving forward and inverse problems based on
  kolmogorov--arnold networks, Computer Methods in Applied Mechanics and
  Engineering 433 (2025) 117518.

\bibitem{zhang2023artificial}
X.~Zhang, L.~Wang, J.~Helwig, Y.~Luo, C.~Fu, Y.~Xie, M.~Liu, Y.~Lin, Z.~Xu,
  K.~Yan, et~al., Artificial intelligence for science in quantum, atomistic,
  and continuum systems, arXiv preprint arXiv:2307.08423 (2023).

\bibitem{wang2022cenn}
Y.~Wang, J.~Sun, W.~Li, Z.~Lu, Y.~Liu, Cenn: Conservative energy method based
  on neural networks with subdomains for solving variational problems involving
  heterogeneous and complex geometries, Computer Methods in Applied Mechanics
  and Engineering 400 (2022) 115491.

\bibitem{he2021towards}
J.~He, C.~Zhou, X.~Ma, T.~Berg-Kirkpatrick, G.~Neubig, Towards a unified view
  of parameter-efficient transfer learning, arXiv preprint arXiv:2110.04366
  (2021).

\bibitem{radford2019language}
A.~Radford, J.~Wu, R.~Child, D.~Luan, D.~Amodei, I.~Sutskever, et~al., Language
  models are unsupervised multitask learners, OpenAI blog 1~(8) (2019) 9.

\bibitem{patankar1983calculation}
S.~V. Patankar, D.~B. Spalding, A calculation procedure for heat, mass and
  momentum transfer in three-dimensional parabolic flows, in: Numerical
  prediction of flow, heat transfer, turbulence and combustion, Elsevier, 1983,
  pp. 54--73.

\bibitem{ill_gradient}
S.~Wang, Y.~Teng, P.~J. S. J. o. S.~C. Perdikaris, Understanding and mitigating
  gradient flow pathologies in physics-informed neural networks, SIAM Journal
  on Scientific Computing 43~(5) (2021) A3055--A3081.

\bibitem{NTK_PINN}
S.~Wang, H.~Wang, P.~Perdikaris, On the eigenvector bias of fourier feature
  networks: From regression to solving multi-scale pdes with physics-informed
  neural networks, Computer Methods in Applied Mechanics and Engineering 384
  (2021) 113938.
\newblock \href {https://doi.org/10.1016/j.cma.2021.113938}
  {\path{doi:10.1016/j.cma.2021.113938}}.

\bibitem{NTK_to_get_hyperparameter_of_PINN}
S.~Wang, X.~Yu, P.~Perdikaris, When and why pinns fail to train: A neural
  tangent kernel perspective, Journal of Computational Physics 449 (2022)
  110768.

\bibitem{zeiler2014visualizing}
M.~D. Zeiler, R.~Fergus, Visualizing and understanding convolutional networks,
  in: Computer Vision--ECCV 2014: 13th European Conference, Zurich,
  Switzerland, September 6-12, 2014, Proceedings, Part I 13, Springer, 2014,
  pp. 818--833.

\bibitem{eshaghi2025applications}
M.~S. Eshaghi, M.~Bamdad, C.~Anitescu, Y.~Wang, X.~Zhuang, T.~Rabczuk,
  Applications of scientific machine learning for the analysis of functionally
  graded porous beams, Neurocomputing 619 (2025) 129119.

\bibitem{PINN_hyperelasticity}
V.~M. Nguyen-Thanh, X.~Zhuang, T.~Rabczuk, A deep energy method for finite
  deformation hyperelasticity, European Journal of Mechanics-A/Solids 80 (2020)
  103874.

\bibitem{wang2023dcm}
Y.~Wang, J.~Sun, T.~Rabczuk, Y.~Liu, Dcem: A deep complementary energy method
  for solid mechanics, International Journal for Numerical Methods in
  Engineering (2024).
\newblock \href {https://doi.org/10.1002/nme.7585}
  {\path{doi:10.1002/nme.7585}}.

\bibitem{automatic_differential}
A.~Paszke, S.~Gross, S.~Chintala, G.~Chanan, E.~Yang, Z.~DeVito, Z.~Lin,
  A.~Desmaison, L.~Antiga, A.~Lerer, Automatic differentiation in pytorch
  (2017).

\end{thebibliography}

\end{document}